\documentclass[10pt,journal,compsoc]{IEEEtran}

\ifCLASSOPTIONcompsoc
\usepackage[nocompress]{cite}
\else
\usepackage{cite}
\fi

\usepackage{amsmath,amsthm,amssymb}
\usepackage{graphicx}
\usepackage{textcomp}
\usepackage{wrapfig}
\usepackage{subfig}
\usepackage{color}
\usepackage{xspace}
\usepackage{overpic}
\usepackage{subfig}
\usepackage{enumitem}
\usepackage{booktabs}
\usepackage{multirow}
\usepackage{color}
\definecolor{blue}{rgb}{0,0,1}
\definecolor{red}{rgb}{1,0,0}
\definecolor{green}{rgb}{0,.5,0}
\definecolor{orange}{rgb}{0.75, 0.4, 0}

\newcommand{\shortcite}{\cite}

\usepackage[colorlinks,linkcolor=red]{hyperref}

\begin{document}

\title{Consistent Two-Flow Network for Tele-Registration of Point Clouds}

\author{
	Zihao~Yan, Zimu~Yi, Ruizhen~Hu, Niloy~J.~Mitra, Daniel~Cohen-Or, and~Hui~Huang
	\IEEEcompsocitemizethanks{
		\IEEEcompsocthanksitem Zihao~Yan, Zimu~Yi, Ruizhen~Hu, and Hui~Huang are with Visual Computing Research Center, College of Computer Science and Software Engineering, Shenzhen University. E-mail: \{mr.salingo, yizimu, ruizhen.hu, hhzhiyan\}@gmail.com
		\IEEEcompsocthanksitem Niloy~J.~Mitra is with University College London and Adobe Research. E-mail: n.mitra@cs.ucl.ac.uk, nimitra@adobe.com
		\IEEEcompsocthanksitem Daniel~Cohen-Or is with Shenzhen University and Tel Aviv University. E-mail: cohenor@gmail.com
		\IEEEcompsocthanksitem Ruizhen~Hu is the corresponding author of this paper
		\IEEEcompsocthanksitem Our code is available at \url{https://github.com/Salingo/CTF-Net}
	}%
}

\IEEEtitleabstractindextext{
	\begin{abstract}
		Rigid registration of partial observations is a fundamental problem in various applied fields. In computer graphics, special attention has been given to the registration between two partial point clouds generated by scanning devices. State-of-the-art registration techniques still struggle when the overlap region between the two point clouds is small, and completely fail if there is no overlap between the scan pairs. In this paper, we present a learning-based technique that alleviates this problem, and allows registration between point clouds, presented in arbitrary poses, and having little or even no overlap, a setting that has been referred to as \emph{tele-registration}. Our technique is based on a novel neural network design that learns a prior of a class of shapes and can complete a partial shape. The key idea is combining the registration and completion tasks in a way that reinforces each other. In particular, we simultaneously train the registration network and completion network  using two coupled flows, one that \emph{register-and-complete}, and one that \emph{complete-and-register}, and encourage the two flows to produce a consistent result. We show that, compared with each separate flow, this two-flow training leads to robust and reliable tele-registration, and hence to a better point cloud prediction that completes the registered scans. It is also worth mentioning that each of the components in our neural network outperforms state-of-the-art methods in both completion and registration. We further analyze our network with several ablation studies and demonstrate its performance on a large number of partial point clouds, both synthetic and real-world, that have only small or no overlap.

	\end{abstract}
	
	}

\maketitle
\IEEEdisplaynontitleabstractindextext
\IEEEpeerreviewmaketitle

\begin{figure*}
	\centering
	\includegraphics[width=0.95\linewidth]{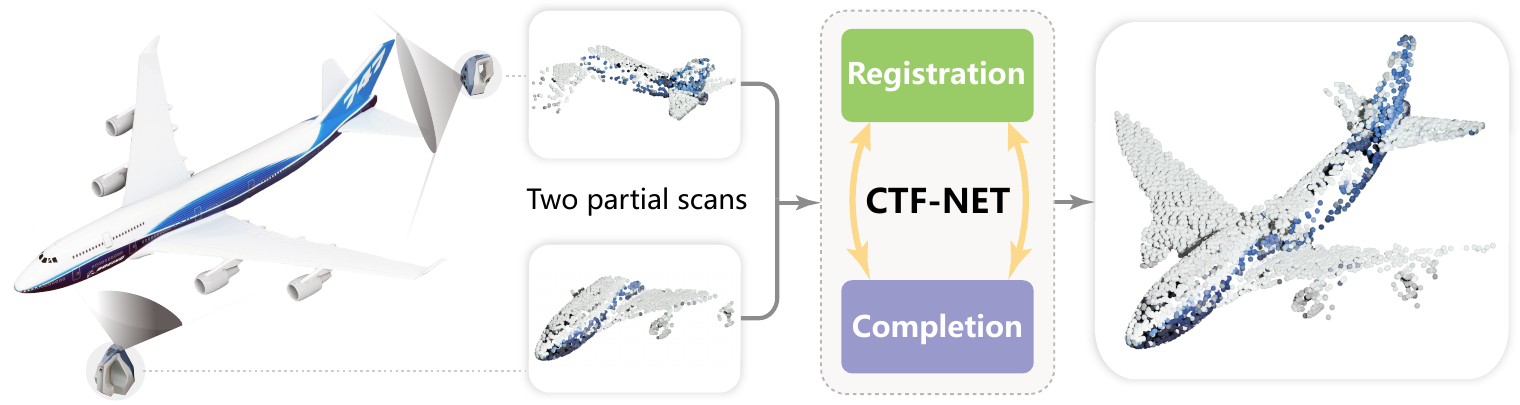}
	\caption{CTF-Net registers pairs of partial scans with little or no overlap. 
			The network is designed to encourage the registration and completion network branches to mutually cooperate to be consistent, and thereby regularizes both the (global)  registration and completion problems. Here, we show the result of the combined registration and completion of two partial scans, with little overlap, of a real scan of a toy airplane.}
	\label{fig:teaser}
\end{figure*}

\section{Introduction} \label{sec:intro}

Shape registration is a long-standing problem with a large variety of methods proposed over the last decades. The registration of partial shapes is significantly more challenging than complete shapes, particularly when the overlap between the parts is small. Popular methods, like RANSAC that match between three or four points, perform well when the overlapping base is large~\cite{fischler1981random, chen1999ransac, rusinkiewicz2001efficient, aiger20084pcs, segal2009generalized}, but completely fail when there is no overlap between the two partial shapes. In the case of partial matching, the information theoretic explanation is that the lack of sufficient information in the scans leads to a family of plausible completions, which in turn results in failure of traditional rigid registration. 

In recent years, neural networks for geometry processing have rapidly emerged and changed the landscape of 3D processing. One notable competence of neural networks is their ability to learn priors of a family of shapes, thus effectively capturing a distribution over possible shapes. Two examples are the reconstruction of a complete shape from a partial input, and the registration of two non-overlapping partial shapes.

We present a rigid registration technique for two partial scans  presented in arbitrary initial poses and having little or no overlap at all. This non-overlapping setting has been referred to as \textit{tele-registration}, and had been attempted in 2D~\cite{Huang2013} and 3D~\cite{Huang2012}, based on a prescribed feature-conforming prior. In our work, we approach the tele-registration problem using learning tools, in particular, deep learning to encode shape priors. The idea is to jointly train two separate networks on the two tasks of shape completion and shape registration of non-overlapping partial shapes.
In training, the networks learn proper priors that allow performing well on these two difficult tasks, rather than treating the tasks independently. 

Our key observation is that registered shapes are easier to be completed than each one alone, and complete shapes are easier to be registered, since their overlap clearly increases. Hence, we combine the registration and completion tasks in a way that reinforces each other. 
In particular, we train the registration and completion networks simultaneously using two coupled flows. One network performs \emph{register-and-complete} and the other \emph{complete-and-register}, such that both registration and completion consistencies are maximized. Fig.~\ref{fig:teaser} illustrates our consistent two-flow network (CTF-Net).
Note that our completion network only generates information for the missing part, and hence the completions along the two branches of the network can be different, and require a dedicated consistency term to produce canonical completion. 
	
Given two partial point cloud inputs with little or no overlap, our method transforms the partial shapes to canonical positions by learning the prior geometry of its class of shapes,
and thus improves the state-of-the-art in terms of completion results on 3D model from the learned class. By comparing our two-flow method to each single-flow and other baselines, we validate that composing two flows together effectively strengthens each component. We evaluate our method on synthetic and real-world examples (e.g., RedWood and Pix3D), and demonstrate the superiority of the approach compared to state-of-the-art methods in both global registration (4PCS, DCP) and completion networks (TopNet, PF-Net).
In summary, we present a method that addresses the problem of non-overlapping point cloud registration, based on a symmetric neural network that is designed to jointly perform registration and completion, in a way that reinforces each other to establish a new state-of-the-art.

\section{Related Work} \label{sec:related}

There has been extensive research on general registration problems and in three dimensions in particular. ICP~\cite{besl1992method} is a widely used algorithm for 3D registration. Some following ICP variants~\cite{rusinkiewicz2001efficient, segal2009generalized, bouaziz2013sparse} aim at improving from different aspects. Recently, DCP~\cite{wang2019dcp} is proposed that revisits ICP from a deep learning perspective.

There is an abundance of SLAM works that deals with registrations and pose estimation at the scene level. For a broad survey, see \cite{tam2012registration}. An apparently similar idea to our is presented by Yang et al.~\shortcite{yang2019extreme}. Unlike our one-step technique, they refine the registration and completion modules iteratively. Yang et al.~\cite{yang2020extreme} also propose hybrid representations for relative pose estimation. These methods, however, match 3D RGB-D scenes rather than point clouds, thus require more information such as color, 360-image.
Chen et al.~\shortcite{chen2019plade} introduce a plane-based descriptor for the point cloud registration with a small overlap. However, for many shapes on object-level, like cars or lamps, it is hard to find a plane surface for matching.
Brachmann et al.~\shortcite{brachmann2017dsac} propose a learning-based method for pose estimation, however, this method is mainly designed for camera localization and it's hard to directly adapt it for shape registration.

In the following, we discuss previous works that are most related to the specific tasks of paired shape registration and partial shape completion on the object level, focusing on deep learning techniques.

\subsection{Paired shape registration} 
Recently, there have been research efforts to apply deep neural networks for the task of rigid~\cite{su2015render} and non-rigid~\cite{hanocka2018alignet, groueix2019unsupervised} registration, to offer faster and more robust than classic techniques.
Elbaz et al. ~\shortcite{elbaz20173d} propose a method that focuses on localizing the close-proximity scanned point cloud in a large-scale point cloud scene. They use super-points to match the corresponding region, and a deep neural network to calculate the transformation between the local and global point cloud.
Yew et al. ~\shortcite{yew20183dfeat} propose a weakly supervised deep learning framework to holistically learn a 3D feature detector and descriptor from GPS/INS tagged 3D point clouds. They use a Siamese architecture that learns to recognize if the given point clouds are from the same location. The correspondences between point clouds are obtained by a learned descriptor vector.
Choy et al. ~\shortcite{choy2020deep} propose a framework for pairwise registration of real-world 3D scans. This method contains a 6-dimensional convolutional network for correspondence confidence prediction, and then the pose is estimated and further refined recursively.

\begin{figure*}[!t]
\centering
\includegraphics[width=0.97\linewidth]{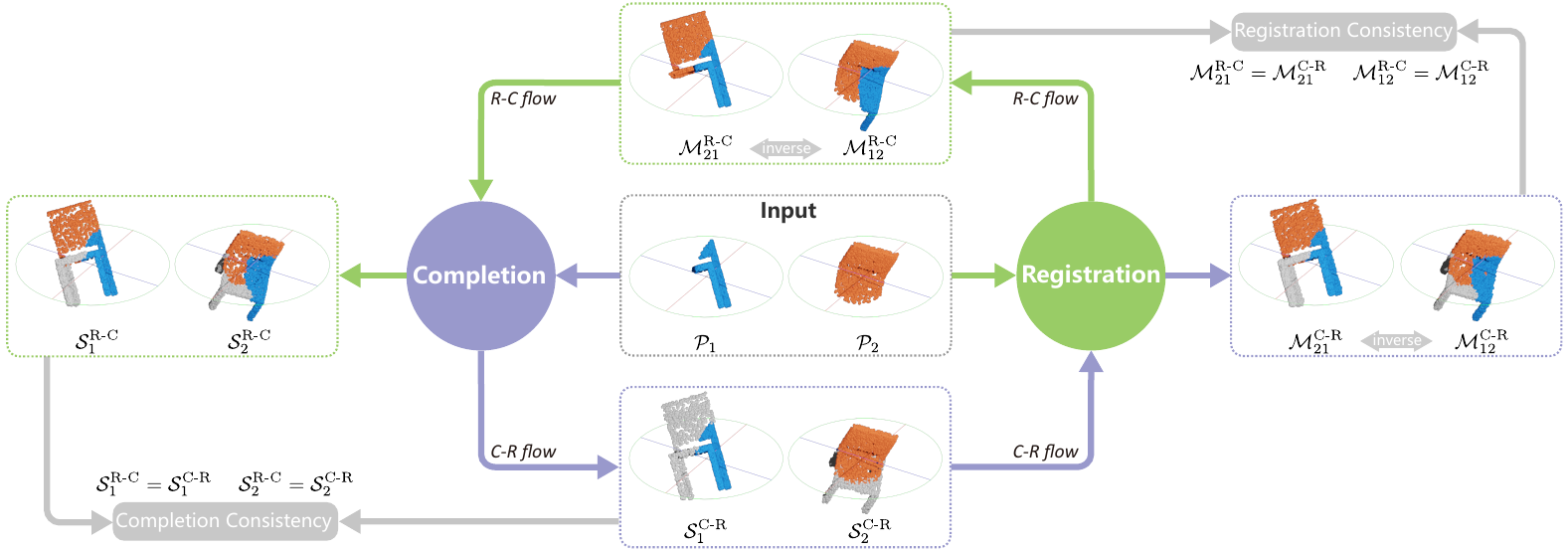}
\caption{The architecture of CTF-Net. Given a pair of partial scans, CTF-Net simultaneously predicts the transformation parameters for registration and coordinates of points for completion. The prediction follows a mirrored manner, which performs \textit{register-and-complete(R-C)} in one flow and \textit{complete-and-register(C-R)} in another. The \textit{R-C flow} and \textit{C-R flow} are denoted by green and purple lines respectively. These two flows mutually reinforce each other by enhancing consistency on their outputs, which is represented by gray lines.}
\label{fig:net_overall} 
\end{figure*}

Aoki et al. ~\shortcite{yaoki2019pointnetlk} propose PointNetLK, a modification to the classical LK(Lucas \& Kanade) algorithm which circumvents the need for convolution on the PointNet representation. This framework for rigid registration is more robust to initialization and missing parts than classic ICP.
Wang et al. ~\shortcite{wang2019dcp} propose Deep Closest Point(DCP), that revisits ICP from a deep learning perspective. The ICP-style method consists of three parts, that learns the common features of the input point clouds to register them together.
Similarly, Yew and Lee \shortcite{yew2020rpm} propose RPM-Net, which is less sensitive to initialization alignment comparing to the original ICP method. However, RPM-Net assumes that the normal information is given in the point cloud data.
Wong and Solomon \shortcite{wang2019prnet} propose PRNet, a sequential decision-making framework to achieve point cloud registration iteratively. Unlike the methods mentioned above, this method is able to handle partial-to-partial registration, the key is to use a detector to find the points in common between partial views, and keypoint-to-keypoint correspondences. These deep-learning based methods are aiming at detecting key points in the input paired shape, then match and pair them to compute the alignment transformation. 
These methods assume that the two parts have a significant overlap region that contains a few key points. 
Huang et al.~\shortcite{Huang2012} present a field-guided algorithm that is able to automatically compose the 3D shape given several pieces of it, where the input pieces have no overlapping. In our work, we utilize a data-driven approach to enable the registration, without the prescribed feature prior. Given two partial point clouds with little or no overlapping, we do registration and completion at the same time, to allow the network to better align the input shapes.

\subsection{Partial shape completion}
There are an increasing number of works focusing on the partial to complete shape generation, many of which are applied on point cloud representation, since it is strongly related to realistic scenarios, where the point clouds are the raw data coming from 3D acquisition devices. PointNet~\cite{qi2017pointnet} proposed a deep learning method for point cloud shape, which promotes several learning-based applications on the point cloud, including completion. Yuan et al.~\shortcite{yuan2018pcn} proposed a method that generates point clouds in two stages, where the first stage is to use a fully-connected decoder to obtain a coarse resolution point cloud and the second stage generates the final output by a folding-based decoder.
Tchapmi et al.~\shortcite{tchapmi2019topnet} introduced a novel decoder for point cloud completion which generates arbitrarily structured point clouds without explicitly enforcing a specific structure. The proposed decoder generates point clouds according to a tree structure where each node of the tree represents a subset of the point cloud.
Wang et al.~\shortcite{wang2020cascaded} proposed a completion method that contains an up-sampling module that predicts denser results than other completion methods.
Huang et al.~\shortcite{huang2020pf} introduced a method that only completes the missing regions of the input shape and can preserve its details, and successfully addressed the blurring issues caused by the auto-encoder structure. However, they still require the input partial data to cover a significant portion of the surface region of the shape, while in many real cases only a small region of the shape is captured by a single scan.

\section{Overview} \label{sec:overview}

Our tele-registration method consists of a two-flow network, as illustrated in Fig.~\ref{fig:net_overall}. The idea is to simultaneously learn two networks, one for registration (colored in green in the figure) and one for completion (colored in purple in the figure). Taking a pair of partial shapes as input, the \textit{C-R flow} branch first completes each partial shape, separately, and then registers the parts, which now have higher overlap, to produce an aligned shape. In the \textit{R-C flow} branch, the input pair is first registered by the registration network, and then completed by the completion network. %

The key of our method is to connect and couple the \textit{R-C} and \textit{C-R} flows with two losses: one is a \textit{registration consistency} loss, which encourages the registration networks in the two flows to predict the same transformation parameters; the other is the \textit{completion consistency} loss, which encourages the two flows to output similar reconstruction results.
As we shall show, the two flows strengthen each other. It should be noted that the complexity of the completion and registration in two flows are different. For the completion network, the one trained in \textit{R-C flow} is easier, since the input shape is already registered, and contains more overlapping geometric information than the one in \textit{C-R flow}. Similarly, the task for the registration network in the \textit{C-R flow} is easier because the input pair shapes are more complete than the one in \textit{R-C flow}. Thus, our final output is the shape completion results from the \textit{R-C flow}, and the registration parameters from the \textit{C-R flow}.

In the following method section, we first describe the input and output of our method, then introduce the registration and completion networks separately, which form the proposed CTF-Net. Finally, we describe the loss functions that enable the two flows to reinforce each other.

\begin{figure*}[!h]
\centering
\includegraphics[width=0.97\linewidth]{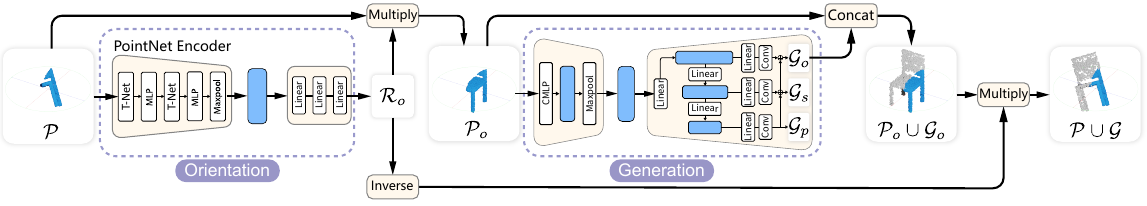}
\caption{The architecture of the completion network. For each partial shape $\mathcal{P}$, the orientation module first predicts the parameter $\mathcal{R}_o$ that rotates $\mathcal{P}$ into the canonical view and obtains $\mathcal{P}_o$. The following generation module predicts the missing part $\mathcal{G}_o$ relative to $\mathcal{P}_o$ by an encoder-decoder pair, and the completed shape $\mathcal{P}_o \cup \mathcal{G}_o$ is then rotated back to the original pose, by multiplying the inverse of $\mathcal{R}_o$, to obtain the final output $\mathcal{S} = \mathcal{P} \cup \mathcal{G}$.}
\label{fig:net_comp}
\end{figure*}

\section{Method} \label{sec:method}

\subsection{Input and output}
The input of our method is a pair of partial point clouds, denoted as $\mathcal{P}_1$ and $\mathcal{P}_2$,  with little overlap  extracted from the same shape $\mathcal{S}$.  The point cloud pairs are centered at the origin point and randomly rotated in 3D. We achieve the goal of tele-registering the point cloud pairs and complete the missing part by making use of our key idea to combine the registration and completion tasks in a way that reinforces each other.
Our outputs include the completion of those two parts in their original states denoted respectively as $\mathcal{S}_1$ and $\mathcal{S}_2$, and the relative transformation between them with $\mathcal{M}_{21}$ denoting the transformation that registers $\mathcal{P}_2 \rightarrow \mathcal{P}_1$ and $\mathcal{M}_{12}$ denoting the transformation that registers $\mathcal{P}_1 \rightarrow \mathcal{P}_2$. 

CTF-Net consists of two flows: \textit{C-R flow} and \textit{R-C flow}. Since each flow provides a set of outputs, we use superscripts to distinguish the two sets of outputs. 

In the \textit{C-R flow},  $\mathcal{P}_1$ and $\mathcal{P}_2$ are first passed through the completion network to get the completion results $\mathcal{S}_1^{\text{C-R }} = \text{C}(\mathcal{P}_1)$ and  $\mathcal{S}_2^{\text{C-R }} = \text{C}(\mathcal{P}_2)$. Subsequently, the two completions are passed to the registration network to output their relative transformation $\mathcal{M}_{12}^{\text{C-R }}$ and $\mathcal{M}_{21}^{\text{C-R }}$. 
In the \textit{R-C flow}, $\mathcal{P}_1$ and $\mathcal{P}_2$, which can come in arbitrary poses, are first passed through the registration network to get their relative transformations $\mathcal{M}_{12}^{\text{R-C }}$ and $\mathcal{M}_{21}^{\text{R-C}}$. The aligned parts are first combined, before passing to the completion network to get the final results $\mathcal{S}_1^{\text{R-C}} = \text{C}(\mathcal{P}_1 \cup \mathcal{M}_{21}^{\text{R-C}} \mathcal{P}_2)$ and $\mathcal{S}_2^{\text{R-C}} = \text{C}(\mathcal{P}_2 \cup \mathcal{M}_{12}^{\text{R-C}} \mathcal{P}_1)$.
	
Our goal is to make sure that the completion and registration outputs of both flows are close to the ground-truth, denoted as  $\mathcal{S}_1^{*}$, $\mathcal{S}_2^{*}$, $\mathcal{M}_1^{*}$ and $\mathcal{M}_2^{*}$, and more importantly, making the two flows mutually consistent.

\subsection{Network Architecture}

\subsubsection{Completion Network}

The completion network, as shown in Fig.~\ref{fig:net_comp}, takes a partial shape $\mathcal{P}$ as input and outputs the completed shape $\mathcal{S} = \text{C}(\mathcal{P})$. Note that similar to~\cite{huang2020pf}, for the given partial shape $\mathcal{P}$, we only generate the missing part $\mathcal{G}$ and thus the final completed shape is a union of those two  $\mathcal{S} =\mathcal{P} \cup \mathcal{G} $. Moreover, different from most of the previous works on completion which assume that the input shapes are all well-aligned, our input partial shape $\mathcal{P}$ are given in an arbitrary orientation. Therefore, our completion network is composed of two modules: one orientation module and one generation module, where the orientation module rotates $\mathcal{P}$ into a canonical pose to facilitate the following generation module and then the generation module generates the missing part to complete the shape. 

Specifically, the orientation module takes the partial shape $\mathcal{P}$ as input, and outputs the rotation transformation $\mathcal{R}_o$, which is then applied to $\mathcal{P}$ to obtain the oriented shape $\mathcal{P}_o$. 
The generation module is adapted from the PF-Net proposed in~\cite{huang2020pf}. The original PF-Net assumes that the input partial shape covers a large portion of the whole shape, thus the number of the generated points is far less than the input.
However, in our work, since the input is usually a much smaller part of the shape, we set the output point number to equal to the input point number. Note that directly modify the output points of the original PF-Net will highly increase the number of network parameters and lead to huge memory cost and slow training speed, so we modify the parameters of each layer to enlarge the output size gradually.
In more detail, we first pass $\mathcal{P}_o$ to a encoder that extracts the feature of dimension $n_c=1920$, which is then passed to a decoder that generates the missing part in three levels: primary $\mathcal{G}_p$, secondary $\mathcal{G}_s$ and the final detailed output $\mathcal{G}_o$, with the points number of 128, 512, 2048, respectively. $\mathcal{G}_o$ is then concatenated with $\mathcal{P}_o$ to form the complete shape. Finally, we multiply $\mathcal{G}_o \cup \mathcal{P}_o$ with the inverse of $\mathcal{R}_o$ to obtain the final complete shape $\mathcal{S}= \text{C}(\mathcal{P}) = \mathcal{G} \cup \mathcal{P} $.

\begin{figure*}[!h]
  \centering
  \includegraphics[width=0.97\linewidth]{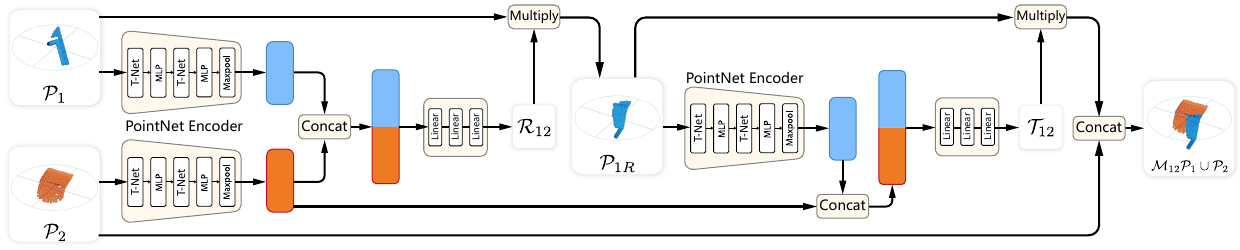}
  \caption{The architecture of the registration network. Taking two paired shapes $\mathcal{P}_{1}$ and $\mathcal{P}_{2}$ and considering $\mathcal{P}_{2}$ as the anchor, the registration network first outputs parameters of a relative rotation $\mathcal{R}_{12}$ from $\mathcal{P}_{1}$ to $\mathcal{P}_{2}$, then, a relative translation $\mathcal{T}_{12}$ from $\mathcal{R}_{12}\mathcal{P}_{1}$ to $\mathcal{P}_{2}$ is predicted. The complete transformation is then denoted as $\mathcal{M}_{12} = \mathcal{T}_{12}\mathcal{R}_{12}$.}
  \label{fig:net_regi}
\end{figure*}

\subsubsection{Registration Network}

The registration network, as shown in Fig.~\ref{fig:net_regi}, takes two shapes, $\mathcal{P}_1$ and $\mathcal{P}_2$, either complete or partial, and outputs the relative transformation $\mathcal{M}_{12}$ from $\mathcal{P}_1 \rightarrow \mathcal{P}_2$ by taking $\mathcal{P}_2$ as the anchor. 	
We decompose the transformation $\mathcal{M}_{12}$ into rotation $\mathcal{R}_{12}$ and translation $\mathcal{T}_{12}$ to reduce the complexity of 3D transformation, i.e., the network first rotates $\mathcal{P}_1$ to make it have the same pose of $\mathcal{P}_2$, and then translates it to align with $\mathcal{P}_2$.
 
Specifically, we first pass the input pair through a PointNet~\cite{qi2017pointnet} encoder, to obtain the feature vector of dimension $n_r=512$ for $\mathcal{P}_1$ and $\mathcal{P}_2$, respectively. The feature vectors are concatenated and passed to a decoder composed of several linear layers, which provides the quaternion parameters that represent the 3D rotation. The quaternion is then converted to rotation matrix $\mathcal{R}_{12}$ and multiplied with $\mathcal{P}_1$ to obtain the rotated part $\mathcal{P}_{1R}$. Note that all the quaternions are normalized in each process. Afterwards, $\mathcal{P}_{1R}$ is passed to the PointNet encoder again to obtain the feature vector of dimension $\mathbb{R}^{n_r}$ ($n_r=512$). The feature is then concatenated again with that extracted from $\mathcal{P}_{2}$ and passed to another set of linear layers to output the set of translation parameters $(dx, dy, dz)$, which are then converted to the translation matrix $\mathcal{T}_{12}$ and multiplied with $\mathcal{P}_{1R}$ to obtain the final registered part. The final transformation matrix is calculated by $\mathcal{M}_{12} = \mathcal{T}_{12} * \mathcal{R}_{12}$.

In both flows, our registration network also takes those two shapes with the other order and output the relative transformation $\mathcal{M}_{21}$ from $\mathcal{P}_2$ to $\mathcal{P}_1$.
Furthermore, in the \textit{C-R flow}, the registration network takes two completed shapes $\mathcal{S}_1$ and $\mathcal{S}_2$ as input. We further apply the transformation on the moving shape, say $\mathcal{S}_1$, and combine it with the anchor shape $\mathcal{S}_2$ to get the registered full shapes. 
Note that since the points size of the registered point cloud is twice that of each input partial point cloud, we down-sample it to half of its original points size in order to keep all the point clouds in the flows to have the same size.

\subsection{Loss Functions}
To train this two-flow network, we design suitable loss functions to account for each type of output and also the consistency between two flows. We define the loss function of our CTF-network as:
\begin{equation}
L = L_{\text{c}} + L_{\text{r}} + L_{\text{s}}.
\end{equation}
where $L_{\text{c}}$ and $L_{\text{r}}$ are the completion and registration loss against the ground-truth, and $L_{\text{s}}$ is the loss to ensure the consistency between two flows.%

\subsubsection{Completion loss.}

Note that our completion network works for input parts with arbitrary orientation, and we rotate the parts into the canonical view first and then generate the missing part to complete the shape. Therefore, the completion loss is computed for outputs from the \textit{R-C} and \textit{C-R} flows against the respective ground truth missing part geometry and rotation transformation as $L_{\text{c}} = \omega_c^{\text{c-r}}L_{\text{c}}^{\text{C-R}} + \omega_o^{\text{c-r}}L_{\text{o}}^{\text{C-R}} + \omega_c^{\text{r-c}}L_{\text{c}}^{\text{R-C}} + \omega_o^{\text{r-c}}L_{\text{o}}^{\text{R-C}} $ with 
\begin{align}
L_{\text{c}}^{\text{C-R}} &= \left(\text{D}_{emd}(\mathcal{G}_1^{\text{C-R}} , \mathcal{G}_1^{\text{C-R}*}) + \text{D}_{emd}(\mathcal{G}_2^{\text{C-R}} , \mathcal{G}_2^{\text{C-R}*})\right) / 2, 
\notag \\ 
L_{\text{c}}^{\text{R-C}} &= \left(\text{D}_{emd}(\mathcal{G}_1^{\text{R-C}} , \mathcal{G}_1^{\text{R-C}*} ) + \text{D}_{emd}(\mathcal{G}_2^{\text{R-C}}, \mathcal{G}_2^{\text{R-C}*})\right) / 2,
\notag \\ 
L_{\text{o}}^{\text{C-R}} &= \left(\text{D}_r(\mathcal{R}_{1o}^{\text{C-R}} , \mathcal{R}_{1o}^{*}) + \text{D}_r(\mathcal{R}_{2o}^{\text{C-R}} , \mathcal{R}_{2o}^{*})\right) / 2,
\notag \\ 
L_{\text{o}}^{\text{R-C}} &= \left(\text{D}_r(\mathcal{R}_{1o}^{\text{R-C}} , \mathcal{R}_{1o}^{*}) + \text{D}_r(\mathcal{R}_{2o}^{\text{R-C}} , \mathcal{R}_{2o}^{*})\right) / 2,
\end{align}
where the weights $\omega_c^{\text{c-r}}$, $\omega_o^{\text{c-r}}$, $\omega_c^{\text{r-c}}$, $\omega_o^{\text{r-c}}$ are set as 1, 3, 0.5, and 1.5, respectively. $\text{D}_{emd}$ is the distance measure between the generated part and the corresponding ground truth, which is defined as the mean of the earth mover's distance (EMD)~\cite{liu2020morphing} computed for all three generated levels($\mathcal{G}_p$, $\mathcal{G}_s$, $\mathcal{G}_o$), and $\text{D}_r$ is the distance measure between two rotations. In more detail, the rotation matrix is converted to a quaternion $\mathcal{Q} = q(\mathcal{R})$. 
Since a quaternion $\mathcal{Q}$ is equivalent to its minus $-\mathcal{Q}$ when representing a rotational transformation, we measure the distance as follows :
\begin{align}
\text{D}_r(\mathcal{R}_1, \mathcal{R}_2) &= \text{D}_q(q(\mathcal{R}_1), q(\mathcal{R}_2)),
\notag \\ 
\text{D}_q(\mathcal{Q}_1, \mathcal{Q}_2) &= min(norm(\mathcal{Q}_1 - \mathcal{Q}_2), norm(\mathcal{Q}_1 + \mathcal{Q}_2)).
\end{align}

Note that for the same part, for example $\mathcal{P}_1$, the ground truth missing part is different in two flows. In the \textit{R-C flow}, $\mathcal{P}_1$ will first be registered and combined with $\mathcal{P}_2$ before completion, so the missing part would be smaller than the one in \textit{C-R flow}. All the ground truth missing parts are extracted and subsampled from the corresponding ground truth complete shape $\mathcal{S}^*$.

\subsubsection{Registration loss.}
The registration loss is also computed for outputs from the \textit{C-R} and \textit{R-C} flows against the respective ground truth transformations as $L_{\text{r}} = \omega_r^{\text{c-r}}L_{\text{r}}^{\text{C-R}} + \omega_r^{\text{r-c}}L_{\text{r}}^{\text{R-C}}$ with 
\begin{align}
L_{\text{r}}^{\text{C-R}} &= \left(\text{D}_m(\mathcal{M}_{12}^{\text{C-R}}, \mathcal{M}_{12}^{*}) + \text{D}_m(\mathcal{M}_{21}^{\text{C-R}}, \mathcal{M}_{21}^{* })\right) / 2,
\notag \\ 
L_{\text{r}}^{\text{R-C}} &= \left(\text{D}_m(\mathcal{M}_{12}^{\text{R-C}}, \mathcal{M}_{12}^{*}) + \text{D}_m(\mathcal{M}_{21}^{\text{R-C}}, \mathcal{M}_{21}^{*})\right) / 2.
\end{align}
where $\text{D}_m$ is the distance measure between two transformations, which is defined as the $\text{D}_q$ between their quaternions plus the mean square error between the translations in the x, z, y axis. The weights $\omega_r^{\text{c-r}}$ and $\omega_r^{\text{r-c}}$ are set as 3 and 9, respectively.

\begin{figure*}[!t]
  \centering
  \includegraphics[width=0.97\linewidth]{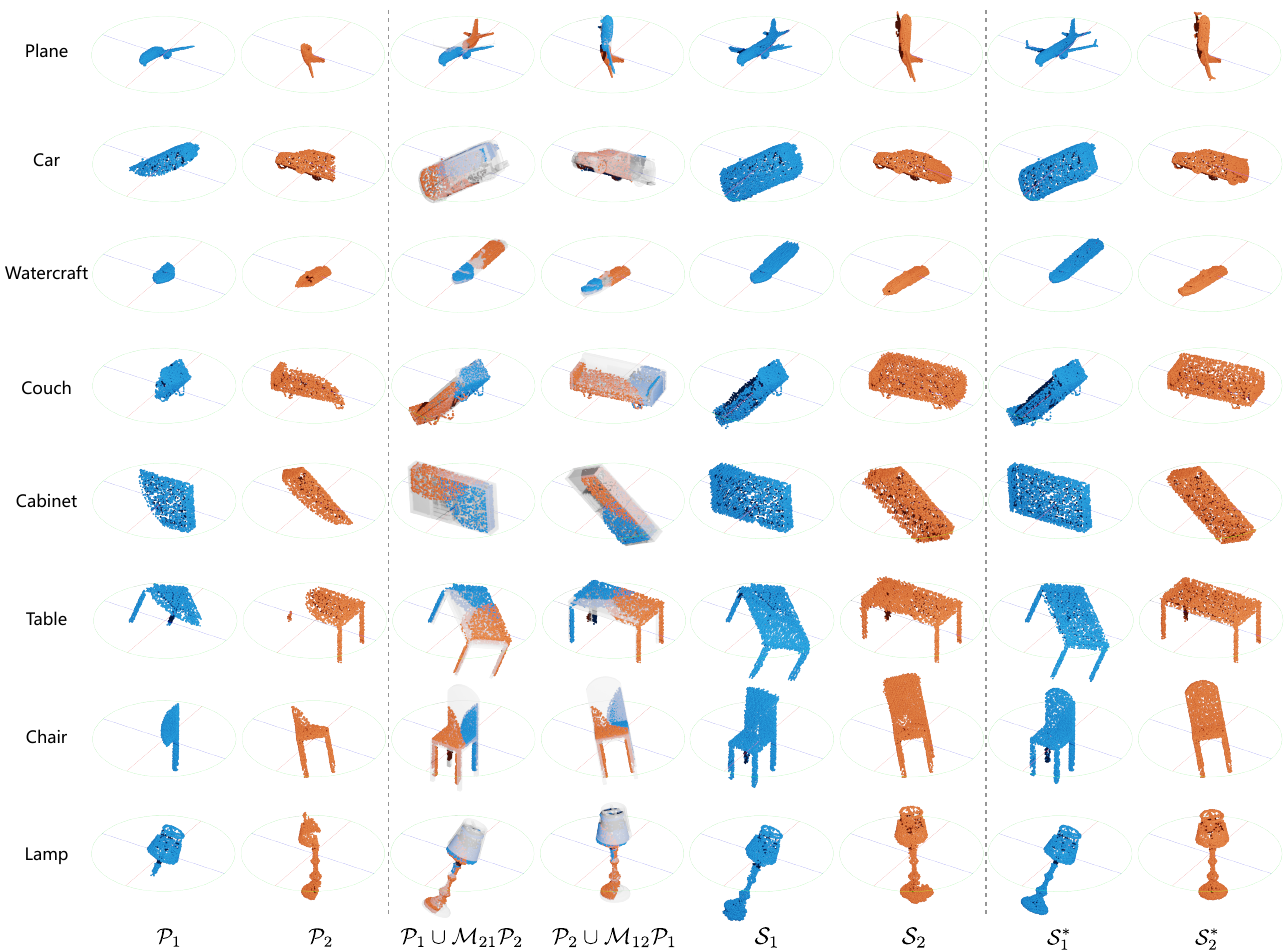}
  \caption{The prediction results of CTF-Net. The first two columns show the input pairs (colored in blue and orange). The third and fourth columns show the registration results from \textit{C-R flow}, where the corresponding original mesh is shown in transparent gray to better display the relative position of the registered parts. The fifth and sixth columns show the completion results from \textit{R-C flow}, and the last two columns are the ground truth point clouds.}
  \label{fig:gallery}
\end{figure*}

\subsubsection{Consistency loss.}
The consistency loss consists of four components:
\begin{equation}
L_{\text{s}} = \omega_{\text{so}} L_{\text{s}}^{\text{O}} + \omega_{\text{sc}} L_{\text{s}}^{\text{C}}+ \omega_{\text{sr}} L_{\text{s}}^{\text{R}} + \omega_{\text{st}} L_{\text{s}}^{\text{T}}.
\end{equation}
where $L_{\text{s}}^{\text{O}}$, $L_{\text{s}}^{\text{C}}$ and $L_{\text{s}}^{\text{R}}$ are the consistency loss defined on the orientation correction, completion and registration in those two flows, while $L_{\text{s}}^{\text{T}}$ is the consistency loss defined on the relative transformations between two shapes with either one as the anchor. In more detail:
\begin{align}
L_{\text{s}}^{\text{O}} &= \left(\text{D}_r(\mathcal{R}_{1o}^{\text{C-R}}, \mathcal{R}_{1o}^{\text{R-C}}) + \text{D}_r(\mathcal{R}_{2o}^{\text{C-R}}, \mathcal{R}_{2o}^{\text{R-C}}) \right) / 2,
\notag \\ 
L_{\text{s}}^{\text{C}} &= \left(\text{D}_{emd}(\mathcal{S}_1^{\text{C-R}}, \mathcal{S}_1^{\text{R-C}}) + \text{D}_{emd}(\mathcal{S}_2^{\text{C-R}}, \mathcal{S}_2^{\text{R-C}}) \right) / 2,
\notag \\ 
L_{\text{s}}^{\text{R}} &= \left(\text{D}_r(\mathcal{M}_{12}^{\text{C-R}}, \mathcal{M}_{12}^{\text{R-C}}) + \text{D}_r(\mathcal{M}_{21}^{\text{C-R}}, \mathcal{M}_{21}^{\text{R-C}}) \right) / 2,
\notag \\ 
L_{\text{s}}^{\text{T}} &= \left(\text{D}_r(\mathcal{M}_{12}^{\text{C-R}}\mathcal{M}_{21}^{\text{C-R}}, I) + \text{D}_r(\mathcal{M}_{12}^{\text{R-C}}\mathcal{M}_{21}^{\text{R-C}}, I) \right) / 2.
\end{align}
where $I$ is the $4\times4$ identity matrix. The weights $\omega_{\text{so}} $, $\omega_{\text{sc}} $, $\omega_{\text{sr}}$ and $\omega_{\text{st}}$ are set as to 3, 1, 3 and 3, respectively by default. For the details about how we choose this set of weights, please refer to the supplementary material.

\section{Results and Evaluation} \label{sec:results}
\begin{figure*}[!t]
  \centering
  \includegraphics[width=0.98\linewidth]{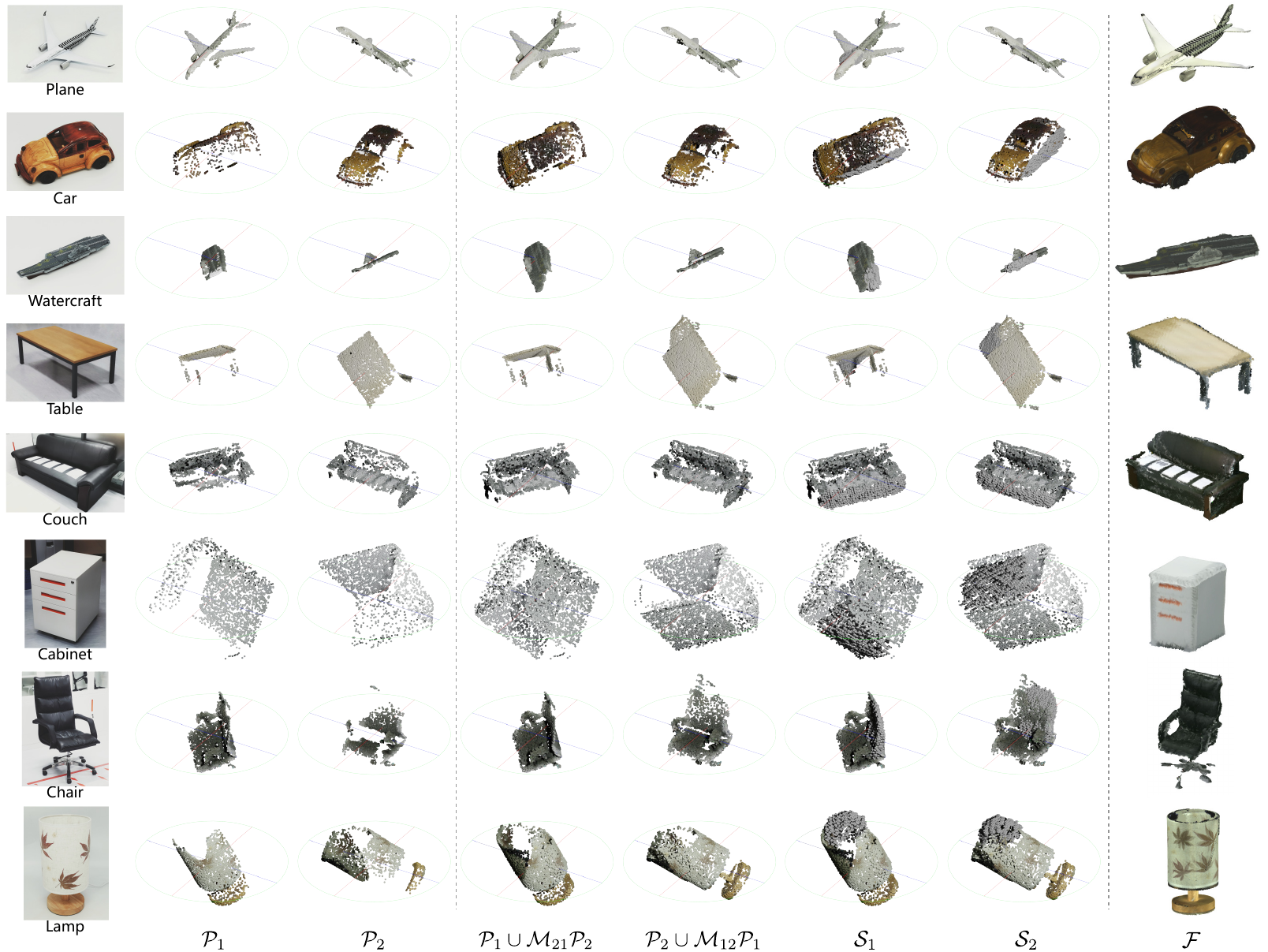}
  \caption{The prediction results on real scans. The first column shows the photo of each real object. The next two columns show the paired partial inputs. The fourth and fifth columns are the registered results, which take each part as an anchor. The fifth and sixth columns show the completion results. The complete fused shape from a much denser RGB-D sequences are shown in the last column for comparison.}
  \label{fig:realscan}
\end{figure*}

\begin{figure*}[!t]
  \centering
  \includegraphics[width=0.98\linewidth]{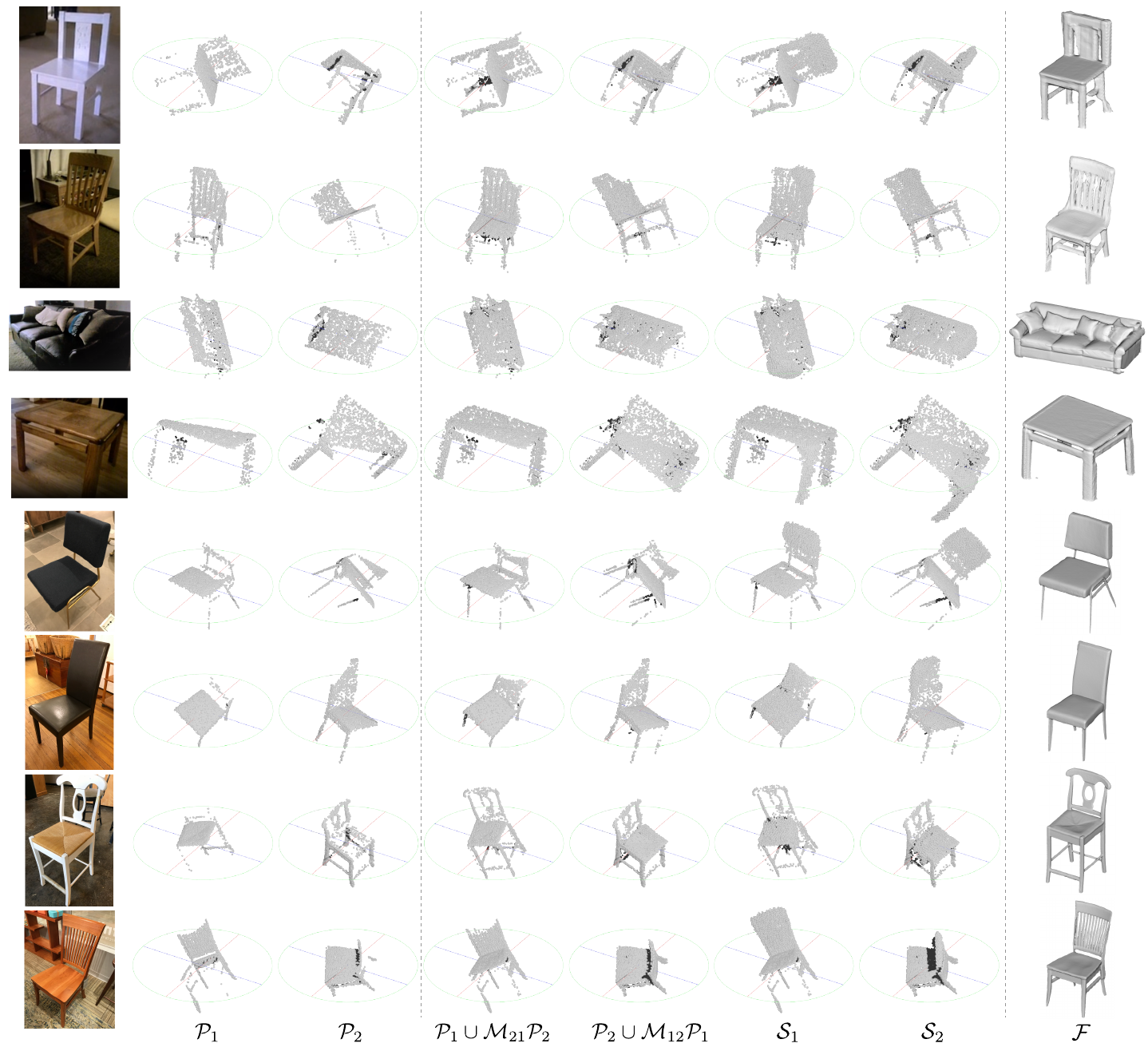}
  \caption{The prediction results on RedWood (upper 4 rows) and Pix3D (lower 4 rows) datasets. The first column shows the photo of each real model, and the next two columns show the paired partial inputs. The fourth and fifth columns are the registered results with either part as the anchor, and the completion results are shown in the following two columns. The last column shows the corresponding reconstructed meshes provided by these two datasets.}
  \label{fig:realscan_pix3d}
\end{figure*}

We first introduce the details of our training data generation in Section~\ref{sec:data}, then, we show results of pair registration and partial completion obtained with our CTF-Net and present a set of qualitative results to demonstrate the capabilities of our method in Section~\ref{sec:qualitative}.
A quantitative evaluation and ablation studies are described in Sections~\ref{sec:quantitative} and~\ref{sec:ablation}. We present results on both synthetic and real data.

\subsection{Data preparation}
\label{sec:data}

Our dataset contains $31,742$ shapes in eight categories from ShapeNet v2~\shortcite{chang2015shapenet}. The training, validating and testing split is similar to the dataset of Tchapmi et al.~\shortcite{tchapmi2019topnet}. Each shape in our dataset is represented as a point cloud which contains $16,384$ points, and all the shapes are normalized into a unit cube centered at the origin.

For each shape $\mathcal{S}^*$ in the dataset, we crop two parts from it and apply random transformations to get the training pairs $(\mathcal{P}_1, \mathcal{P}_2)$. Specifically, we first randomly generate two spheres centered at the surface of the bounding sphere of each shape, with the radius sampled from $[0.3,1.3]$, and then use these two spheres to crop two parts out of the shape. We use this method to simulate the real scan cases when a camera is held and scan around the shape. Note that we set a minimum distance ($0.3$ in our setting) between each pair of cropped parts to avoid too much overlap between those two parts, where the distance is calculated by the Euclidean distance between the centers of those two parts. We also ensure that each cropped part and the remaining part both have more than $N=4096$ points, so that they can be down-sampled to $N=2048$ points.
Each part is first translated to the origin and then rotated about its center. 

The rotation and translation matrix for pairs of parts are denoted as $\mathcal{R}_{1}$, $\mathcal{T}_{1}$, $\mathcal{R}_{2}$ and $\mathcal{T}_{2}$ respectively, and the complete transformations are then denoted as $\mathcal{M}_{1} = \mathcal{R}_{1} \mathcal{T}_{1}$ and $\mathcal{M}_{2} = \mathcal{R}_{2} \mathcal{T}_{2}$. 
Therefore, the corresponding ground truth for completed shapes are $\mathcal{S}_{1}^* = \mathcal{M}_{1} \mathcal{S}^*$ and $\mathcal{S}_{2}^* = \mathcal{M}_{2} \mathcal{S}^*$,
and the corresponding ground truth for relative transformations between those two parts are $\mathcal{M}_{21}^{*} = \mathcal{M}_1\mathcal{M}_2^{-1}$ and $\mathcal{M}_{12}^{*} = \mathcal{M}_2\mathcal{M}_1^{-1}$.
Note that in the following experiments, CTF-Net is trained individually for each of those eight categories. For a fair comparison, all the compared methods are trained category-by-category.

\subsection{Qualitative results}

 \label{sec:qualitative}
We show results for synthetic shapes on eight categories. To verify the generality of our method, we also test our method on real scans.

\subsubsection{Results on synthetic data}

Fig.~\ref{fig:gallery} shows visual examples of the registration and  completion results. %
The input pair of parts are centered at the origin point, as shown in the first two columns.
By taking either part as the anchor, our method is able to transfer the other part to make it align with the anchor well, as shown in the third and fourth columns respectively. 
We can see that even when there is little overlap between the input parts, like most of the examples in the results, CTF-Net can still register them correctly. 
Also, the union shapes after registration with different anchor parts are quite similar thanks to the effect of $L_{\text{s}}^{\text{T}}$.
The fifth and sixth columns show the completed shape for each part, and the last two columns are the corresponding ground truth complete shape. 
We can see that the completion module in our CTF-Net can work well for parts given in two different orientations, and by solving registration and completion together, the parts do not need to be pre-aligned before completion.
Note that we take registration results from the \textit{C-R flow} and completion results from the \textit{R-C flow}, mainly because each will improve the performance for the other, and the end output of each flow is more reliable even with the consistency loss.

\subsubsection{Results on real data}
To demonstrate the generality of our method, we use 3D scanners to manually scan several objects from all those eight categories for testing. Specifically, for small objects with detailed textures, in particular, \texttt{Plane}, \texttt{Car}, \texttt{Watercraft}, and \texttt{Lamp}, we place each object on a turnplate and use an Artec Spider scanner for scanning which directly produces the reconstructed model; for other larger objects, we use a Microsoft Kinect v2 scanner to do the scanning, then utilize bundle fusion~\cite{dai2017bundle} to obtain the reconstructed model. %
The reconstruction results are shown in the last column in Fig.~\ref{fig:realscan}, denoted as $\mathcal{F}$.

To generate the input pairs, we first normalize the reconstructed full model into a unit box and put it in a virtual environment, and then randomly place two cameras pointing at the model to capture RGB-D images. Two partial point clouds are then obtained by back-projecting the depth images to the 3D reconstructed model, and those two single-view point clouds are taken as input to test our CTF-Net. 
Note that here we do not use the single-view point cloud directly from the scanner since there exist some affections such as reflect light which could make the single-view point cloud extremely sparse.

\begin{table*}[!t]
	\caption{Quantitative results on all eight categories. The registration error is quantified using $E_{\theta}$ and $E_{t}$  while the completion error is quantified using $E_{c}^g$ and $E_{c}^f$. More details about the measures are specified in section~\ref{sec:quantitative}.}
	\label{tab:result}
	\renewcommand\arraystretch{1.2}
	\begin{center}
	\begin{tabular}{c||c|c|c|c|c|c|c|c||c}
		\hline
		              & Plane  & Cabinet & Car   & Chair  & Lamp   & Couch  & Table  & Watercraft & Average\\ \hline
		 $E_{\theta}$ & 9.582  & 18.405  & 8.172 & 15.114 & 26.126 & 11.625 & 15.486 & 16.434 & 15.118 \\ \hline
		 $E_{t}$      & 3.927  & 5.605   & 1.606 & 4.144  & 13.925 & 4.045  & 7.399  & 7.105  & 5.970 \\ \hline
		 $E_{emd}^g$  & 3.573  & 5.792   & 3.041 & 6.274  & 11.573 & 5.142  & 5.725  & 5.580  & 5.838 \\ \hline
		 $E_{emd}^f$  & 1.603  & 2.695   & 1.457 & 2.729  & 3.765  & 2.322  & 2.491  & 2.126  & 2.399 \\ \hline
	\end{tabular}
	\end{center}
\end{table*}

Fig.~\ref{fig:realscan} shows the results on our real scanned data. Note that the input partial parts can still be correctly aligned even the point clouds are noisy and non-uniformly distributed. 
For example, although the two parts of the couch shown on the fifth row are quite noisy and covers different region of the couch, our method is still able to successfully align them and complete the missing region in the front.

To further justify the generality of our method, we also tested our method on RedWood~\cite{choi2016large} and Pix3D~\cite{sun2018pix3d} datasets, both of which provided the full reconstructed model.
Similar to our real scanned data, the input pairs of these two datasets are generated by single-view scanning.
Some example results are shown in Fig.~\ref{fig:realscan_pix3d}. %
For each example, the full reconstructed model is shown in the last column, denote as $\mathcal{F}$, for comparison. 
We see that although the single-view input data is quite different from the synthetic training data, the predicted registration is still quite accurate and the completion results are also quite reasonable.
For example, in the fourth row, our method predicts the missing leg of the table, and in the fifth row, we can notice that the entire chair back is successfully reconstructed.
Note that for each real scan example, we slightly rotated the view during visualization to show the single-view inputs more clearly.

\subsection{Quantitative evaluation} \label{sec:quantitative}

We perform quantitative evaluation of both registration and completion networks by measuring the errors of predicted transformation parameters and reconstructed point clouds.

For registration, we calculate the error for the final output of the \textit{C-R flow}. \textit{Rotational error} $E_{\theta}$ is calculated as the absolute error between the predicted and ground truth angle in degrees, and \textit{Translation error} $E_{\text{t}}$ is calculated by the L2 distance between the predicted and ground truth translation (in normalized units):
\begin{align}
E_{\theta} &= \left(|\theta(\mathcal{R}_{12}^{\text{C-R}}, \mathcal{R}_{12}^{\text{C-R}*})| + |\theta(\mathcal{R}_{21}^{\text{C-R}}, \mathcal{R}_{21}^{\text{C-R}*})|\right) / 2,  \\ 
E_{t} &= \left(|t(\mathcal{T}_{12}^{\text{C-R}}, \mathcal{T}_{12}^{\text{C-R}*})| + |t(\mathcal{T}_{21}^{\text{C-R}}, \mathcal{T}_{21}^{\text{C-R}*})|\right) / 2 \times 10^3. 
\end{align}
where $\theta$ denotes the angle derived from the relative quaternion from the predicted quaternion $\mathcal{R}^{\text{C-R}}$ to the ground truth quaternion $\mathcal{R}^{\text{C-R}*}$, and $t$ denotes the distance between the translation decomposed from the predicted transformation $\mathcal{T}^{\text{C-R}}$ and the ground truth transformation $\mathcal{T}^{\text{C-R}*}$.

For completion, we also calculate the error for the completion obtained from the \textit{C-R flow}.
Since our completion network generates the missing region without modifying the given part, we use two measures $E_{\text{emd}}^g$ and $E_{\text{emd}}^f$ to compute the errors in the generated region and full shape, respectively, where $E_{\text{emd}}^g$ is calculated as the EMD between the reconstructed and corresponding ground truth point cloud of the missing region, and  $E_{\text{emd}}^f$ is calculated as the EMD between the reconstructed shape concatenate with the input part and ground truth full shape:
\begin{equation}
E_{emd}^g = \left(\text{D}_{emd}(\mathcal{G}_{1}^{\text{C-R}}, \mathcal{G}_{1}^{\text{C-R}*}) + \text{D}_{emd}(\mathcal{G}_{2}^{\text{C-R}}, \mathcal{G}_{2}^{\text{C-R}*})\right) / 2 \times 10^3, \label{compeval1}
\end{equation}
\begin{equation}
E_{emd}^f = \left(\text{D}_{emd}(\mathcal{S}_{1}^{\text{C-R}} , \mathcal{S}_{1}^{\text{C-R}*} ) + \text{D}_{emd}(\mathcal{S}_{2}^{\text{C-R}} , \mathcal{S}_{2}^{\text{C-R}*})\right) / 2 \times 10^3. \label{compeval2}
\end{equation}

Table~\ref{tab:result} shows the errors of all eight categories. Note that $E_t$, $E_{emd}^g$ and $E_{emd}^f$ values have been scaled by $10^3$ to amplify the error.
In most categories, our CTF-Net is able to predict accurate registration together with completion.
We observe that the registration error of \texttt{Lamp} is significant higher than other categories due to the ambiguity arising from its strong symmetry. For example, the stands of most lamps are cylindrical, which could lead to the rotational ambiguity. Furthermore, the variety and complexity of \texttt{Lamp} category are very large, e.g., a lamp with multi-fold stand, or a pendant lamp with cluttered accessories.
The rotational error of other categories are all lower than 20 degrees; meanwhile, the full shape completion errors of these categories are lower than $3$, meaning that the good reconstruction quality is achieved.

Overall, we achieve an average of 15.12 degrees rotational and 5.97 translation error in registration, and 2.40 completion error for the whole shape.

\subsubsection{Comparison on registration}

We compare the registration results of our method to five other options:
\begin{enumerate}
	\item A classic registration method 4PCS~\cite{aiger20084pcs}, which can globally register complete or partial point sets;
	\item A deep-learning based registration method PRNet~\cite{wang2019prnet}, which focuses on partial-to-partial registration, with self-supervised learning;
	\item A deep-learning based registration method DCP~\cite{wang2019dcp}, can be seen as state-of-the-art;
	\item Baseline registration network which predicts the rotation and translation parameters at once from a single decoder, denoted as BL-Regi. The detailed network structure can be found in the supplementary material;
	\item Our registration network alone, denoted as Regi.
\end{enumerate}

\begin{figure*}[!t]
  \centering
  \includegraphics[width=0.97\linewidth]{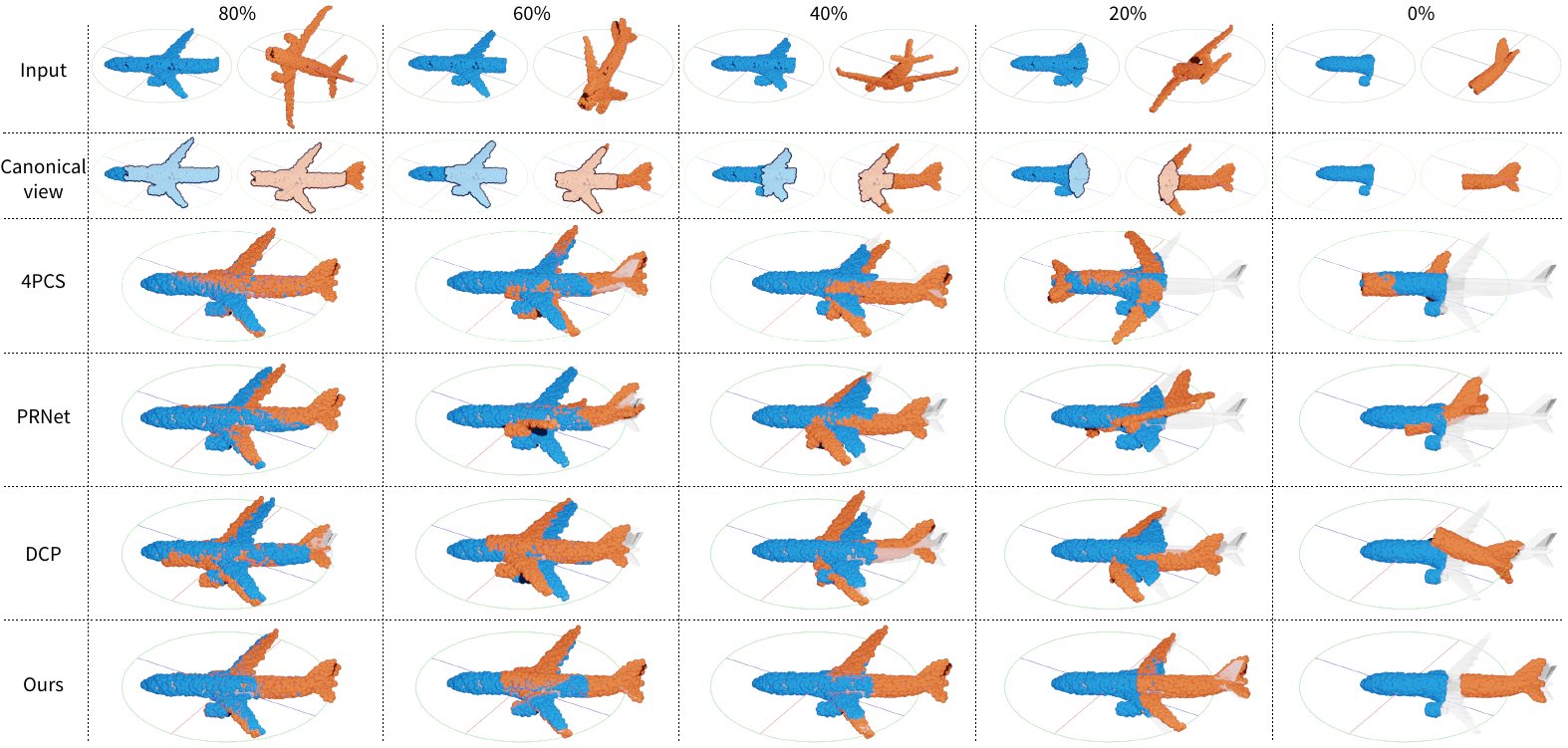}
  \caption{Visual comparisons of our CTF-Net to other registration methods on different overlap data. The input pair of partial parts is shown in the first row. To better illustrate the overlap region, we show the input pair in the second row after rotating each part into a canonical view, and mark the overlap region with a lighter color. Next four rows show the registration results provided by 4PCS, DCP, PRNet and ours respectively. Here we only show the registration results that takes the blue part as the anchor.}
  \label{fig:overlap}
\end{figure*}

\begin{table}[!t]
	\caption{Quantitative comparisons of different registration methods.}
	\label{tab:comparison_regi}
	\renewcommand\arraystretch{1.2}
	\small
	\begin{minipage}{\columnwidth}
		\begin{center}
		\resizebox{\textwidth}{!}{
		\begin{tabular}{c||c|c|c|c|c|c}
			\hline
			              & 4PCS    & PRNet  & DCP    & BL-Regi & Regi   & \textbf{Ours} \\ \hline
			 $E_{\theta}$ & 138.910 & 25.796 & 17.767 & 21.336  & 17.912 & \textbf{15.118} \\ \hline
			 $E_{t}$      & 62.456  & 28.188 & 22.317 & 17.305  & 6.927  & \textbf{5.970} \\ \hline
		\end{tabular}}
		\end{center}
	\end{minipage}
\end{table}

\begin{figure}[!t]
  \centering
  \includegraphics[width=\linewidth]{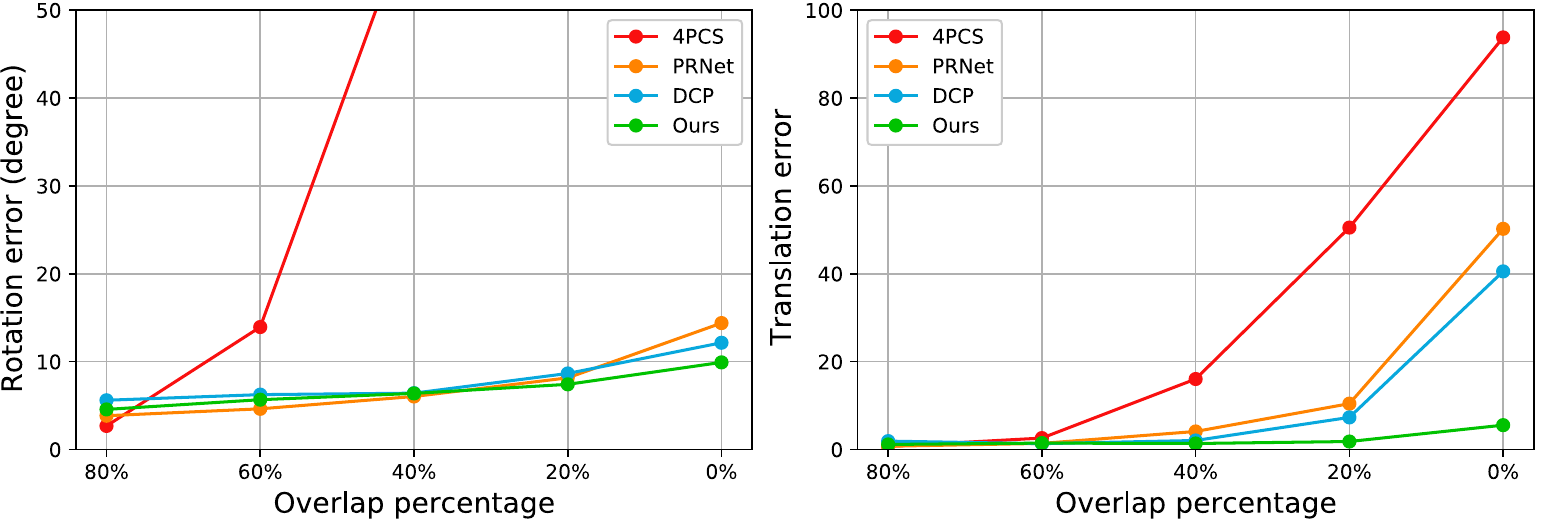}
  \caption{Quantitative comparisons of the rotation and translation errors of our CTF-Net on different overlapping data. In the left chart, we crop the error value larger than 50 degrees to show the difference between 4PCS, PRNet, DCP and our method more clearly.}
  \label{fig:chart_overlap}
\end{figure}

The geometry-based method 4PCS is directly tested on our dataset, while the remaining three learning-based methods are trained/tested on the same dataset as ours. 
The comparison of prediction error is reported in Table~\ref{tab:comparison_regi}.
We can see that the prediction error of 4PCS is the highest since it assumes that there are certain-level overlap between the input pair, while our data are mainly non-overlap. 
The errors obtained using PRNet are lower than 4PCS but higher than others, meaning that PRNet can not works well on our data with little overlap between the input pairs. DCP performs better comparing to our BL-Regi method in rotational error, however, the translation error of BL-Regi is 23\% lower. Comparing DCP to our single registration module Regi, we can see that the rotational error is quite close, but the translation error of Regi is 68\% lower. The key idea of DCP is to find the correspondences between two point sets, which also fails on our dataset.
The baseline registration method, which predicts the rotation and translation parameters simultaneously from the same decoder, has slightly higher rotation error to our registration network, however, the translation error is more than twice as much as ours. This proves that splitting the prediction of rotation and translation could achieve better performance.
Last but not least, our method with two complete flows gets the best results comparing to all five other options, which shows the benefit of combining the registration and completion tasks.

In order to assess the advantage of our method on non-overlap data, we further compare our method with 4PCS, PRNet and DCP on the data with different levels of overlap.
To generate testing pairs with a certain overlap $\eta$, we modify our data preparation procedure slightly. In more detail, for each shape in our testing set, we randomly generate two spheres with fixed centers $(0, 0.75, 0)$ and $(0, -0.75, 0)$, respectively, and the radius randomly sampled from $[0.3,1.3]$ to crop two parts from a complete shape. We then calculate the IoU of the cropped parts and keep the ones with IoU in $[0.9\eta,1.1\eta]$. Finally, we downsample each part to $N=2048$ points.

We take the \texttt{Plane} category for testing, and quantitative comparisons of the rotation and translation errors are shown in Fig.~\ref{fig:chart_overlap}. 
We can see that both the rotation and translation errors of almost all the methods keep increasing as the overlap region decreases.
Regards to rotation error, 4PCS performs best when the overlap region is 80\%, however, as the overlap decreases, the performance of 4PCS dropped significantly and the error is lager than 50 degrees from 40\% overlap to 0\%. 

\begin{figure*}[!t]
  \centering
  \includegraphics[width=0.97\linewidth]{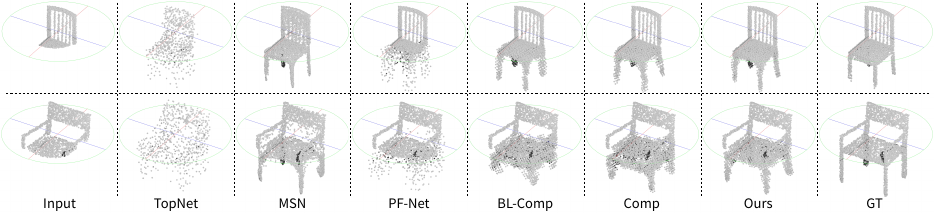}
  \caption{Visual comparisons of results obtained by different completion methods. Note that all the results are rotated to canonical view to facilitate the visual comparison.}
  \label{fig:compcompare}
\end{figure*}

PRNet and DCP performs more stable than 4PCS,  and the rotational error of DCP is slightly lower than ours at 80\% and 60\% overlap, but CTF-Net keeps the lowest error from 40\% to 0\%.
For translation error, the four methods are quite similar at 80\% overlap, however, the error of 4PCS increases rapidly when overlap region decreases. 
The error of DCP is slightly lower than PRNet but significantly higher than ours at 20\% and 0\%.
PRNet is supposed to be able to deal with a partial-to-partial registration, however, it shows worse performance than DCP in our setting. The main reason is that PRNet assumes a large overlap between the input pair data and requires a good initial alignment for the ensuing iteration, while for our input data with quite little overlap, they fail to find a sufficiently good initial alignment.

We observe that the translation error of our method decreases a bit even the overlap region drop from 80\% to 20\%, implying that our network has higher capability in learning data with little overlap.

Fig.~\ref{fig:overlap} shows the visual comparison of our method to 4PCS, PRNet, and DCP.
To highlight the differences in the overlap regions, we rotate all the input pairs and align their first part (shown in blue), and use a lighter color to show the overlap region between the input pair, shown in the second row. %
From the results, we can see that 4PCS fails at 20\% and 0\% overlap, since the algorithm assumes that two parts should have a large overlapped region. For the deep-learning based methods PRNet and DCP, we observe that the rotational error of PRNet is large at and 60\%, 20\% and 0\% overlap. DCP produces more accurate rotation angle, however, the translation error is still high.
Our method performs stable on different overlap data. We observe that even there is no overlap between the head and tail of the plane, our method is still able to align two parts in the correct position, with a certain margin in the middle.
This experiment also shows that our method are robust when the area of the input data variants, which is helpful for the use of real scanned data.

\subsubsection{Comparison on completion}
 \begin{table}[t!]
	\caption{Quantitative comparisons of different completion methods.}
	\label{tab:comparison_comp}
	\renewcommand\arraystretch{1.2}
	\small
	\begin{minipage}{\columnwidth}
		\begin{center}
		\resizebox{\textwidth}{!}{
		\begin{tabular}{c||c|c|c|c|c|c}
			\hline
			             & TopNet & MSN    & PF-Net  & BL-Comp & Comp   & \textbf{Ours} \\ \hline
		     $E_{emd}^g$ & -      & -      & 10.560  & 8.330   & 6.937  & \textbf{5.838} \\ \hline
		     $E_{emd}^f$ & 10.003 & 3.536  & 4.919   & 3.811   & 3.172  & \textbf{2.399} \\ \hline
		     $E_{cd}^g$  & -      & -      & \textbf{7.245}   & 10.290  & 8.183  & 7.339 \\ \hline
		     $E_{cd}^f$  & 2.735  & 2.274  & \textbf{2.187}   & 3.460   & 2.959  & 2.466 \\ \hline
		\end{tabular}}
		\end{center}
	\end{minipage}
\end{table}

We compare the completion results of our method to five other options:
\begin{enumerate}
    \item TopNet~\cite{tchapmi2019topnet}, which direct reconstruct the whole shape using a structural decoder;
    \item MSN~\cite{liu2020morphing}, which predicts the whole shape in a coarse to fine manner;
    \item The original PF-Net~\cite{huang2020pf}, which can be seen as state-of-the-art; %
	\item Our completion network alone without orientation module, denoted as BL-Comp;
	\item Our completion network alone, denoted as Comp;
\end{enumerate}

All the five methods are trained/tested on the same dataset as ours, TopNet and PF-Net are trained using chamfer distance (CD) as proposed in their paper, and the remaining methods are trained using EMD. For fair comparison, we add another two completion quality measures based on CD instead of EMD, denoted as $E_{\text{cd}}^g$ and $E_{\text{cd}}^f$, by substituting $D_{emd}$ by $D_{cd}$ in Equation~\ref{compeval1} and Equation~\ref{compeval2}. Specifically,
\begin{align}
E_{cd}^g &= \left(\text{D}_{cd}(\mathcal{G}_{1}^{\text{C-R}}, \mathcal{G}_{1}^{\text{C-R}*}) + \text{D}_{cd}(\mathcal{G}_{2}^{\text{C-R}}, \mathcal{G}_{2}^{\text{C-R}*})\right) / 2 \times 10^4, \\
E_{cd}^f &= \left(\text{D}_{cd}(\mathcal{S}_{1}^{\text{C-R}} , \mathcal{S}_{1}^{\text{C-R}*} ) + \text{D}_{cd}(\mathcal{S}_{2}^{\text{C-R}} , \mathcal{S}_{2}^{\text{C-R}*})\right) / 2 \times 10^4. 
\end{align}
Note that $E_{\text{cd}}^g$ and $E_{\text{cd}}^f$ values have been scaled by $10^4$ to amplify the error.
The reconstruction error is reported in Table~\ref{tab:comparison_comp}.
TopNet and MSN predict the whole shape directly, thus we only compute $E_{emd}^f$ and $E_{cd}^f$ for them. For all other methods, only points on the missing regions are generated, so $E_{emd}^g$ and $E_{cd}^g$ are also computed. 
We can see that TopNet gets the highest error $E_{emd}^f$ since it doesn't keep the points from the input part and it is trained using CD. 
For comparison, MSN also predicts the whole shape, but it is trained using EMD, so both $E_{emd}^f$ and $E_{cd}^f$ are lower than TopNet. 
PF-Net achieves the lowest error in $E_{cd}^g$ and $E_{cd}^f$, since it keeps the input region and is trained using CD. 
The main difference between BL-Comp and PF-Net is that BL-Comp is trained using EMD measure, thus both $E_{emd}^g$ and $E_{emd}^f$ are lower than PF-Net.
To further improve the results on input part pairs with randomly 3D rotation, we added an orientation module to BL-Comp, denoted as Comp, and all the errors get lower comparing to BL-Comp. %
Our method obtains the best result in EMD measure, with 76\%, 32\%, and 51\% drop on $E_{emd}^f$ comparing to TopNet, MSN and PF-Net, thanks to the consistent two flow network.

Fig.~\ref{fig:compcompare} shows two examples of different completion methods. We see that the prediction results of TopNet are quite blurred and lack fine details. Similar results can be seen in the fourth column, i.e., PF-Net keeps the original input but the predicted parts are sparse, leading to the inconsistent results. These two methods are trained using CD measure, which is not able to effectively constrain the sparseness of the point cloud. The results of MSN are obviously more compact, however, we can see that there are some outlier points distributed in the blank region. For example, there are some unnecessary points  located between the back and the seat region in the second row. Our baseline completion network performs better than PF-Net, but incurs some distortions in the output. The completion method, which includes the orientation module to the baseline completion, predicts points more accurately. Our method achieves the best results comparing to others, and the generated points are evenly distributed in the missing regions. Note that here we take the completion results from \textit{C-R flow} to compare with the other methods.

\begin{figure}[!t]
  \includegraphics[width=0.99\linewidth]{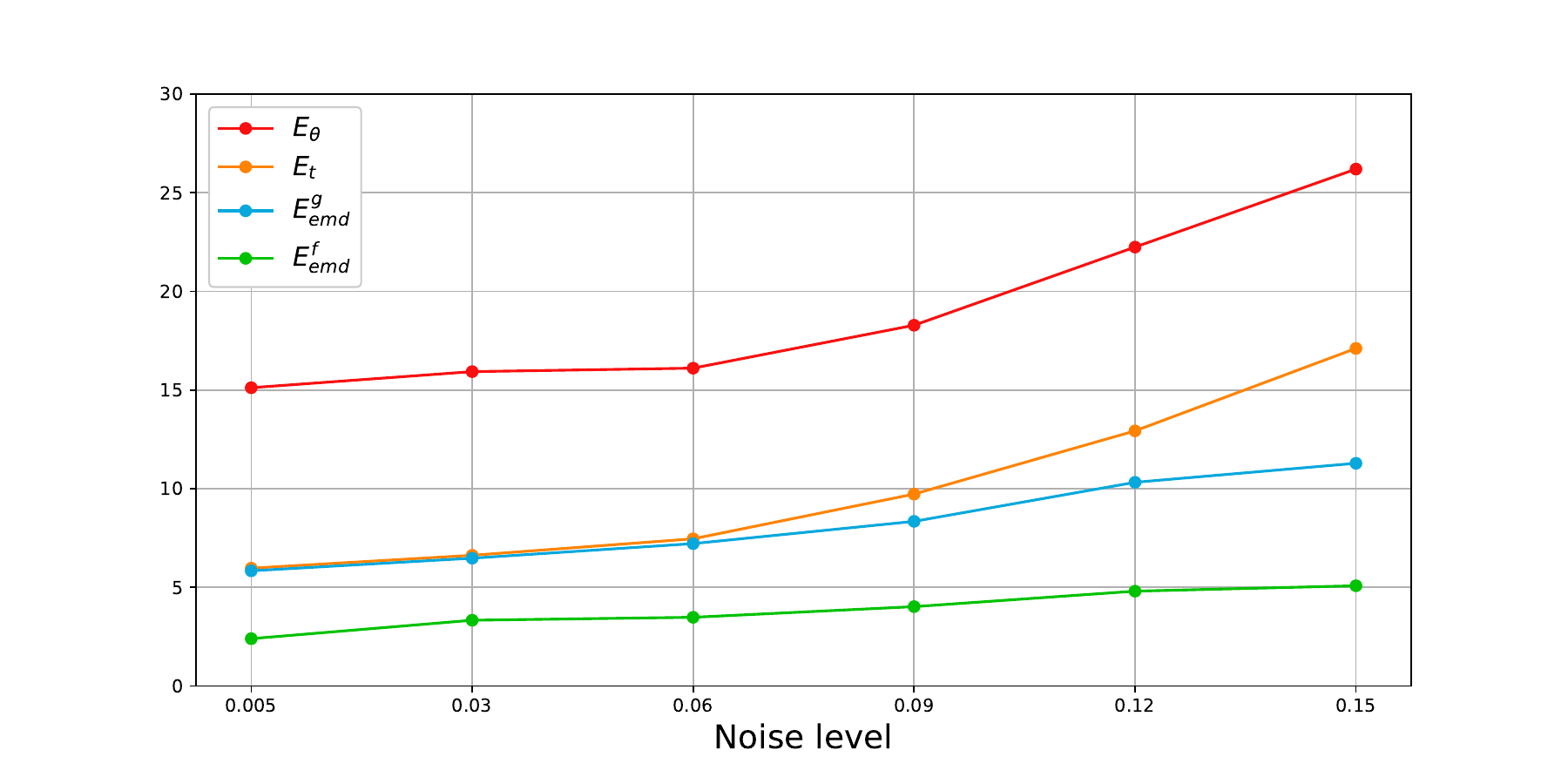}
  \caption{Quantitative evalution of the stress test with the increase of noise level added to the input parts.}
  \label{fig:chart_stress}
\end{figure}

\begin{figure}[!t]
  \centering
  \includegraphics[width=0.9\linewidth]{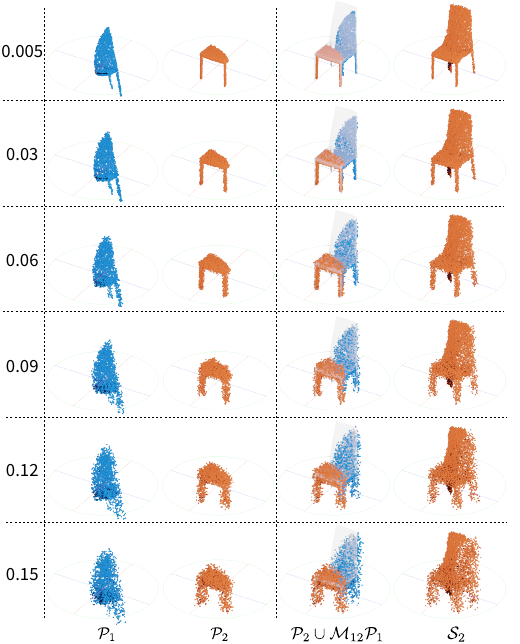}
  \caption{Visual examples of the stress test. Given a pair of input parts with different noise level (first two columns),  the registration (third column) and completion (last column) results are presented.}
  \label{fig:stress}
\end{figure}

\subsubsection{Stress test}
To better simulate the scenario of real scanning, we generate the data with different noise level to test the robustness of our CTF-Net. Specifically, to generate the data with noise level $\zeta$, we randomly sample $N \times 3$ values range in $[0, \zeta]$ as the noise, then add the sampled noise to the original input partial pair $\mathcal{P}_1$ and $\mathcal{P}_2$. We set five different $\zeta$ to test our method, and the results are shown in Fig.~\ref{fig:chart_stress}. Note that our original data generated by virtual scanning have the noise level at 0.005.
We observe that all the errors increase quite slightly as the noise level increases from 0.005 to 0.09, i.e., $E_\theta$, $E_t$, $E_{emd}^g$ and $E_{emd}^f$ keep lower than $20$, $10$, $8$ and $5$ respectively. 
At noise level of 0.12 and 0.15, which rarely appear in real scanning, the errors increasing slightly faster, but the rotational error keeps lower than $27$ degree. 
The results show that our method is able to maintain low prediction error given noisy data.

Fig.~\ref{fig:stress} shows how the registration and completion errors change with the increase of noise level added to the input parts.
The registration of our method is correct even the input partial data are quite noisy, as shown in the examples in the last row. Besides, the missing parts generated by our method are recognizable at all noise levels. 

\begin{figure}[!t]
  \centering
  \includegraphics[width=0.93\linewidth]{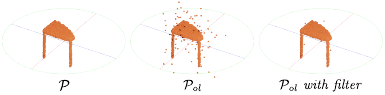}
  \caption{Visual example of the input point cloud with outliers. We show original scan, scan with 100 outliers, scan with 100 outliers processed by a filter, respectively.}
  \label{fig:outlier}
\end{figure}

Other than points with random and small noise, outliers can also exist in the captured point cloud data, thus we perform another experiment to show how outliers affect the performance of our CTF-Net.
Specifically, as shown in Fig.~\ref{fig:outlier}, for each testing point cloud data $\mathcal{P}$, we randomly select $K$ points and each of them is then translated by a random displacement to simulate the outliers. The direction of each displacement is uniformly sampled in $SE(3)$ space, and the length of each displacement is uniformly sampled between $[0.1,0.5]$. In this experiment, we set $K$ to be 50 and 100, respectively, to test our method. We denote the processed data as $\mathcal{P}_{ol}$. 

\begin{figure}[!t]
  \includegraphics[width=0.99\linewidth]{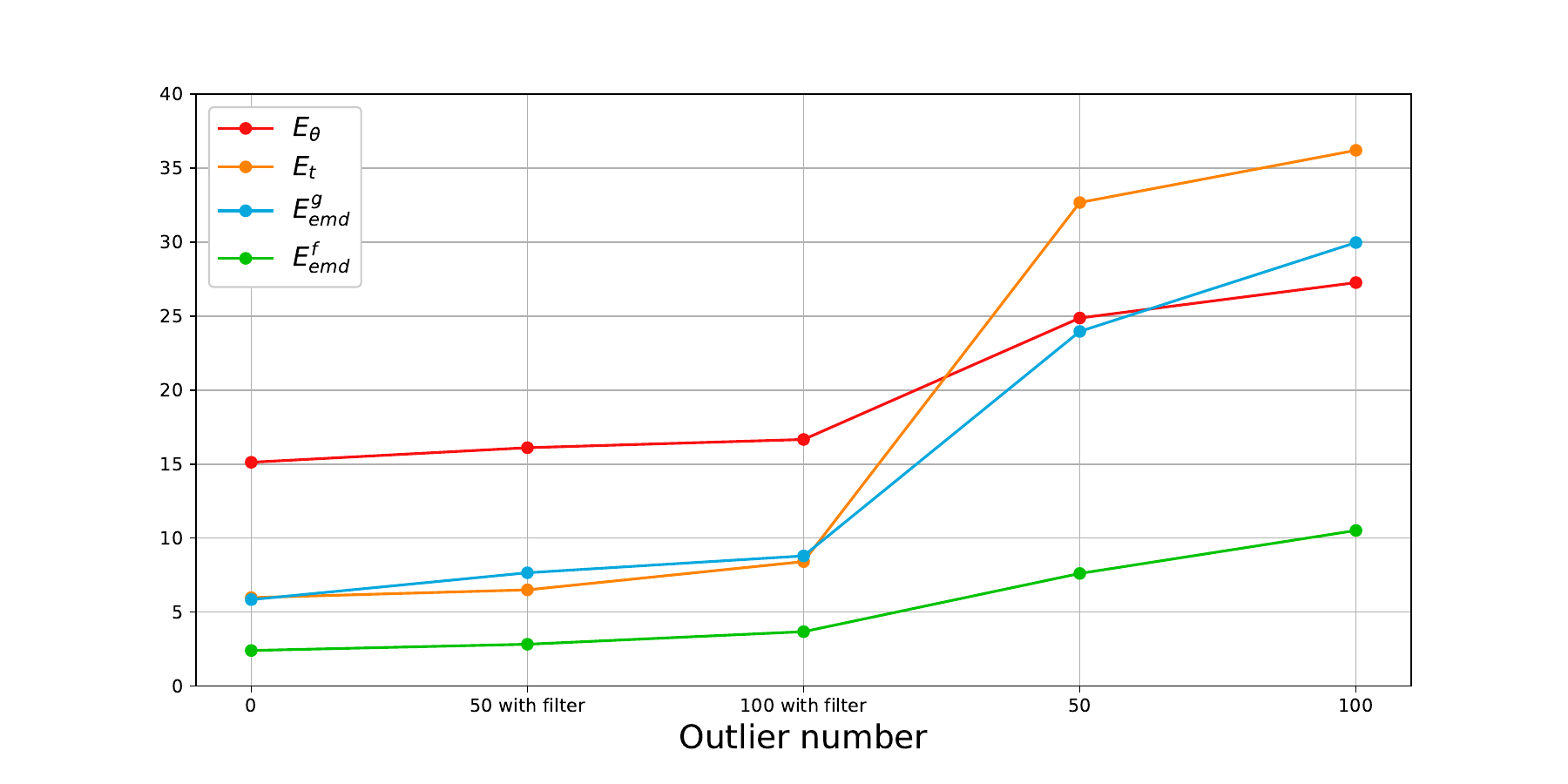}
  \caption{Quantitative evaluation of the stress test with an increasing number of outliers added to the input scans with and without filtering.}
  \label{fig:chart_stress_outlier}
\end{figure}

Furthermore, since automatic outlier removal algorithms are widely used in many works related to point cloud processing, we also test our CTF-Net on the outlier data after being processed by an outlier filter. Specifically, we utilize the radius outlier removal method from Zhou et al.~\shortcite{zhou2018open3d}, which removes points that have few neighbors in a given sphere around them,  and we use the default parameters setting proposed in~\shortcite{zhou2018open3d}. The processed data are denoted as $\mathcal{P}_{ol}\textit{ with filter}$.

The results are reported in Fig.~\ref{fig:chart_stress_outlier}. We observe that the errors are relatively high when the outliers are added to the input scans without filtering, especially for $E_t$. Such amount of random outliers will confuse the network when distinguishing the part poses, thus lead to high errors. However, when we apply a filter to the data contaminated with outliers, the errors remain very close compared to those of the results on clean data, meaning that our method is able to perform well on the data cleaned up by a simple and automatic outlier removal filter.

\subsection{Ablation studies} \label{sec:ablation}

To justify the network structure and loss functions designed in our method, we perform several ablation studies for our CTF-Net.
For network structure, we take single \textit{R-C flow} and single \textit{C-R flow} to perform the comparison, the results are shown in Table~\ref{tab:ablation_net}. Since both \textit{R-C flow} and \textit{C-R flow} can produce the transformation parameters and completed shapes, we use the same measure as described in Section~\ref{sec:quantitative}.
Since our CTF-Net contains twice parameters comparing to each single flow, for fair comparison, we slightly modify \textit{R-C flow} and \textit{C-R flow} to make them have comparable parameters to ours, denoted as $\textit{R-C}^1$ and $\textit{C-R}^1$.
Specifically, we double the output dimension of each layer (except for the last layer) of all the decoders, and then obtain the network with almost twice parameters as the original single flow network.
In addition, the registration network in our CTF-Net is trained with both partial point clouds and (predicted) complete point clouds and the completion network in our CTF-Net is trained with both partial input point clouds and (predicted) registered point clouds, so we compare our method to each single flow which trained with augmented data.
Specifically, for single \textit{R-C flow}, we train the completion network with both input partial point cloud and combined point cloud after registration, denoted as $\textit{R-C}^2$, while for single \textit{C-R flow}, we train the registration network with both input partial point cloud and predicted complete point cloud, denoted as $\textit{C-R}^2$.

From the results, we can see that for single \textit{R-C flow}, the errors of both rotational and translation is significantly higher than our method, which shows that the direct registration of two non-overlap partial shapes is quite challenging. For the version with double parameters $\textit{R-C}^1$, the errors are quite similar to \textit{R-C flow}. For $\textit{R-C}^2$, we observe that training the completion network with additional input can improve the results of completion comparing to single \textit{R-C flow}, but the errors are still higher than our method.
The errors of single \textit{C-R flow} are also higher than ours, the registration error of $\textit{C-R}^1$ is lower than \textit{C-R}, but the completion error increases. For $\textit{C-R}^2$, the error of registration is lower than \textit{C-R} and $\textit{C-R}^1$, however, the errors are higher than CTF-Net.
The experiments show that the performance of either single flow is not comparable to our CTF-Net, even with the doubled parameters,  and only the combination of two flows with consistency loss can actually improve the results.

\begin{table}[!t]
	\caption{Ablation studies in which we compare our CTF-Net to versions where we remove either flow of the CTF-Net.}
	\label{tab:ablation_net}
	\renewcommand\arraystretch{1.2}
	\small
	\begin{minipage}{\columnwidth}
		\begin{center}
		\resizebox{\textwidth}{!}{
		\begin{tabular}{c||c|c|c|c|c|c|c}
			\hline
			 & $\textit{R-C}$ & $\textit{R-C}^1$ & $\textit{R-C}^2$ & $\textit{C-R}$ & $\textit{C-R}^1$ & $\textit{C-R}^2$ & \textbf{Ours} \\ \hline
			 $E_{\theta}$   & 19.138 & 18.777 & 18.844 & 18.383 & 17.565 & 16.974 & {\bf 15.118} \\ \hline
			 $E_{t}$        & 7.919  & 8.187  & 8.124  & 7.296  & 6.349  & 6.033  & {\bf 5.970} \\ \hline
		     $E_{emd}^g$    & 7.495  & 7.230  & 7.041  & 7.421  & 7.633  & 7.587  & {\bf 5.838} \\ \hline
		     $E_{emd}^f$    & 3.640  & 3.594  & 3.478  & 3.652  & 3.952  & 3.869  & {\bf 2.399} \\ \hline
		\end{tabular}}
		\end{center}
	\end{minipage}
\end{table}

\begin{table}[!t]
	\caption{Ablation studies in which we compare our CTF-Net to versions where we remove selected terms of the loss function.}
	\label{tab:ablation_loss}
	\renewcommand\arraystretch{1.2}
	\small
	\begin{minipage}{\columnwidth}
		\begin{center}
		\resizebox{\textwidth}{!}{
		\begin{tabular}{c||c|c|c|c|c}
			\hline
			 & w/o $L_{\text{s}}^{\text{O}}$ & w/o $L_{\text{s}}^{\text{C}}$ & w/o $L_{\text{s}}^{\text{R}}$ & w/o $L_{\text{s}}^{\text{T}}$ & \textbf{Ours} \\ \hline
			 $E_{\theta}$   & 17.038 & 18.767 & 19.014 & 16.345 & {\bf 15.118} \\ \hline
			 $E_{t}$        & 5.995 & 6.864 & 6.053 & 6.290 & {\bf 5.970} \\ \hline
		     $E_{emd}^g$    & 6.965 & 7.928 & 6.824 & 6.393 & {\bf 5.838} \\ \hline
		     $E_{emd}^f$    & 3.180 & 3.876 & 3.044 & 2.561 & {\bf 2.399} \\ \hline
		\end{tabular}}
		\end{center}
	\end{minipage}
\end{table}

We design four consistency losses after combining the \textit{R-C flow} and \textit{C-R flow} together: $L_{\text{s}}^{\text{O}}$, $L_{\text{s}}^{\text{C}}$, $L_{\text{s}}^{\text{R}}$, $L_{\text{s}}^{\text{T}}$.
To show the effectiveness of our consistency losses, we perform comparisons of our method with and without each loss term, as shown in Table~\ref{tab:ablation_loss}. By comparing the errors of the last five columns, we validate that the consistency losses are able to make the two flows strengthen each other, which leads to the lowest error.
We can see that both registration and completion errors are higher without the reconstruction consistency loss. Removing the parameter consistency loss results in the registration between two parts being less consistent, and thus leads to larger errors in prediction.

\section{Discussion and Future Work} \label{sec:future}

We presented CTF-Net for tele-registration of two partial point clouds. Our method excels where the surfaces represented by point clouds have little or no overlap. The success of our method is attributed to the competence of neural network to learn a prior of a class of shapes, which allows predicting complete shapes from partial observations and registering non-overlapping parts. The key architectural design comes in the form of consistency between the \emph{register-and-complete} and the \emph{complete-and-register} networks.

The consistency of the two network flows encourages the two networks to predict reliably and performs surprisingly well. Given that parts $\mathcal{P}_1$ and $\mathcal{P}_2$ are disjoint, then their completions $\mathcal{S}_1$ and $\mathcal{S}_2$ should agree \textit{without} conditioning. We expect $\mathcal{S}_1$ to trivially agree with $\mathcal{P}_1$ in the overlap regions, but the rest of $\mathcal{S}_1$ is ambiguous. Here, we also expect $\mathcal{S}_1$ to agree with $\mathcal{P}_2$ in the overlapping region, which is a hard task.
Thus, the completion task by itself is ambiguous, while the registration with no overlap is ill-posed. Solving both tasks jointly and consistently, however, makes the problems more well-posed and less ambiguous.

\begin{figure}[!t]
  \centering
  \includegraphics[width=0.88\linewidth]{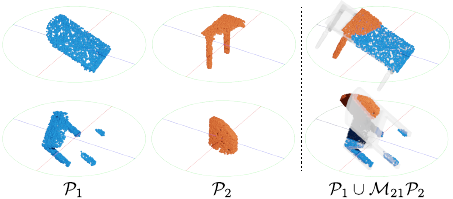}
  \caption{Failure cases of registration prediction with CTF-Net. The upper row contains a table where two parts with some freedom on the degree of overlap, and the lower one contains a chair where the symmetry of one part can cause misplacement of the other part.}
  \label{fig:faliure}
\end{figure}

{\em Limitations} of data-driven methods are naturally carried over to our method. The range by which our method works well is directly derived by the training data and the capacity of the network, which includes the variety of the expected input geometries and their mutual pre-disposition. 
As shown in Fig.~\ref{fig:faliure}, our CTF-Net may fail in registration prediction if the input pairs are sufficiently ambiguous. The upper row shows a table example, where the matching parts have very similar geometry and the ground truth degree of overlap between two input parts are ambiguous, which can cause different registration results. If we assume that the overlap region is larger, the two parts will be placed closer to each other as the result shown in the figure. 
For the chair shown in the second row, since the complete version of $\mathcal{P}_1$ that captures the leg and seat information can be symmetric, so the back part (complete version) of $\mathcal{P}_2$ can be put on any side to make it a valid chair, which introduce the ambiguity of registration between $\mathcal{P}_1$ and $\mathcal{P}_2$. As shown in the figure, the back in placed on the right side instead of back side of the seat.
Another limitation is that our dataset is generated by sphere cropping, since the ground truth for each network is obtained in this way. However, a more natural way to create training data might be by back-projecting depth images to 3D.
Also, currently, our method is trained category-by-category, which may limit the generality. We are planning to enable cross-category training by improving the network structures and loss functions in the future.
Further, our current implementation considers the coordinates of the point cloud, and ignores the additional attributes that can be associated with scanned data, like normals and colors. We leave this for future research.

Another direction is to leverage the tele-registration of our method to align the parts from different objects. For example, given a chair without legs and a partial cabinet, one may register them to form a new object that have both the functionality of chair and cabinet. This part-based modeling requires well designed dataset and network.

In the context of continuous scanning (e.g., using Kinect Fusion), data fusion works better in short sequences than on longer ones, where tracking and registration errors accumulate. 
In our tests, we found CTF-Net to
complement scanning performance by successfully stitching shorter bursts of fused results. This emphasizes the importance of tele-registration for disjoint scans in the context of autonomous (or semi-autonomous) robots and drones.

\section*{Acknowledgment}
This work was supported by NSFC (U2001206, 61872250), GD Talent Program (2019JC05X328), GD Natural Science Foundation (2021B1515020085, 2020A0505100064), DEGP Key Project (2018KZDXM058), Shenzhen Science and Technology Program (RCJC20200714114435012), Royal Society (NAF-R1-180099), ISF (2472/17, 2492/20), and GD Laboratory of Artificial Intelligence and Digital Economy (SZ).

\bibliographystyle{IEEEtran}
\bibliography{IEEEabrv,CTF}

\begin{IEEEbiography}[{\includegraphics[width=1in,height=1.25in,clip,keepaspectratio]{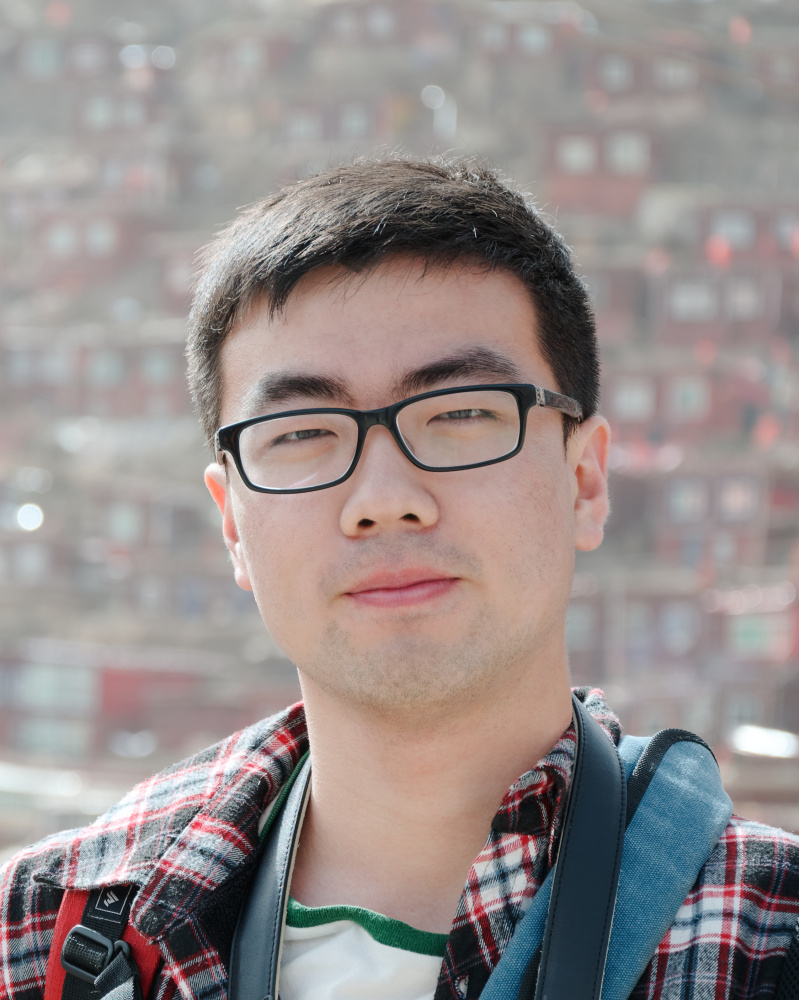}}]{Zihao~Yan}
	is a Ph.D. candidate in the Visual Computing Research Center at Shenzhen University. He received his Bachelor's degree in Electrical Engineering from the University of Electronic Science and Technology of China in 2016. His research interests include shape/scene understanding, point cloud analysis, 3D reconstruction.
\end{IEEEbiography}

\begin{IEEEbiography}[{\includegraphics[width=1in,height=1.25in,clip,keepaspectratio]{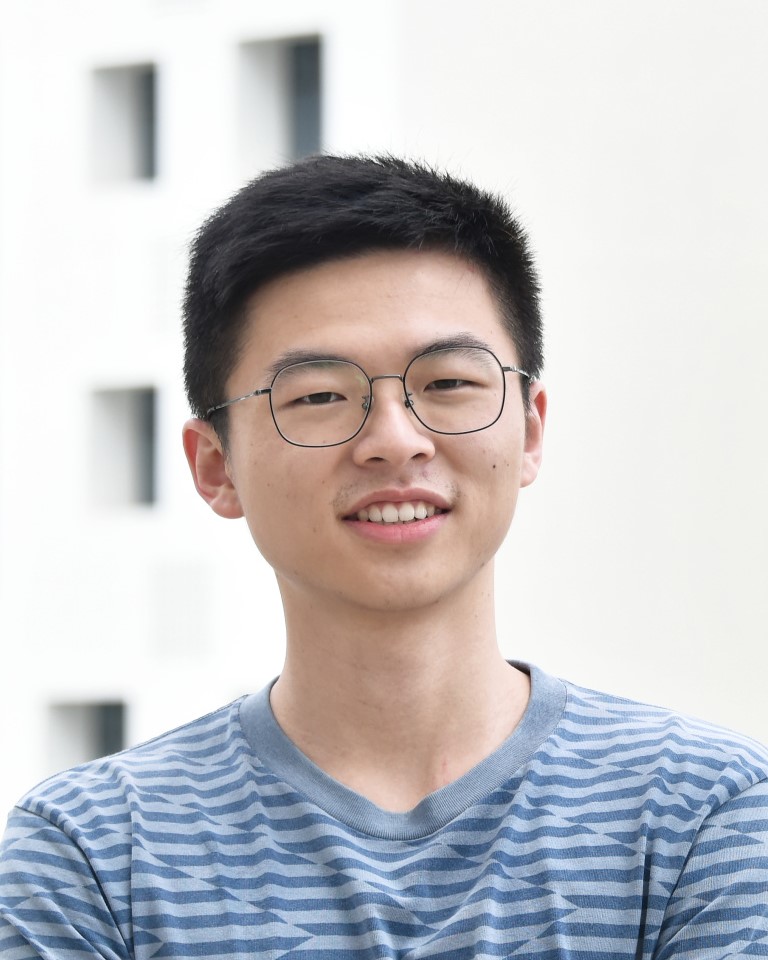}}]{Zimu~Yi}
	is a M.Sc. candidate in the Visual Computing Research Center at Shenzhen University. He received his Bachelor's degree in Computer Science from Beihang University in 2019. His research interest lies in computer graphics.
\end{IEEEbiography}

\begin{IEEEbiography}[{\includegraphics[width=1in,height=1.25in,clip,keepaspectratio]{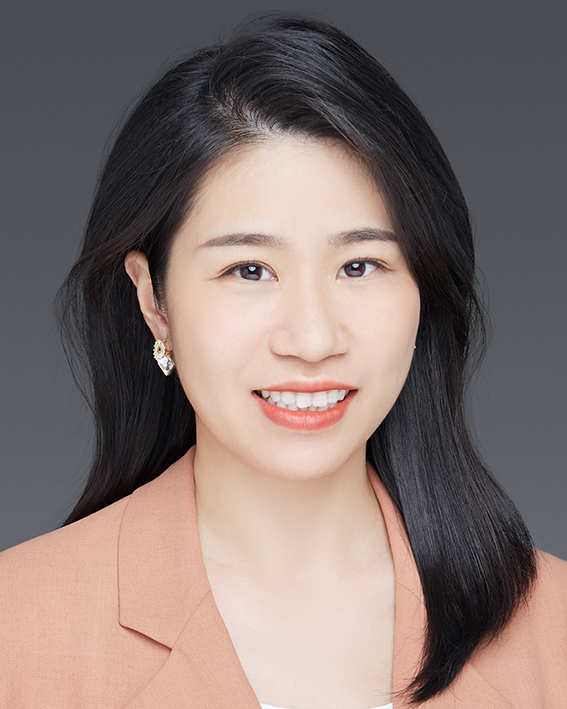}}]{Ruizhen~Hu}
	is an Associate Professor at Shenzhen University, China. She received her Ph.D. from the Department of Mathematics, Zhejiang University. Before that, she spent two years visiting Simon Fraser University, Canada. Her research interests are in computer graphics, with a recent focus on applying machine learning to advance the understanding and generative modeling of visual data including 3D shapes and indoor scenes. She is an editorial board member of The Visual Computer and IEEE CG\&A.
\end{IEEEbiography}

\begin{IEEEbiography}[{\includegraphics[width=1in,height=1.25in,clip,keepaspectratio]{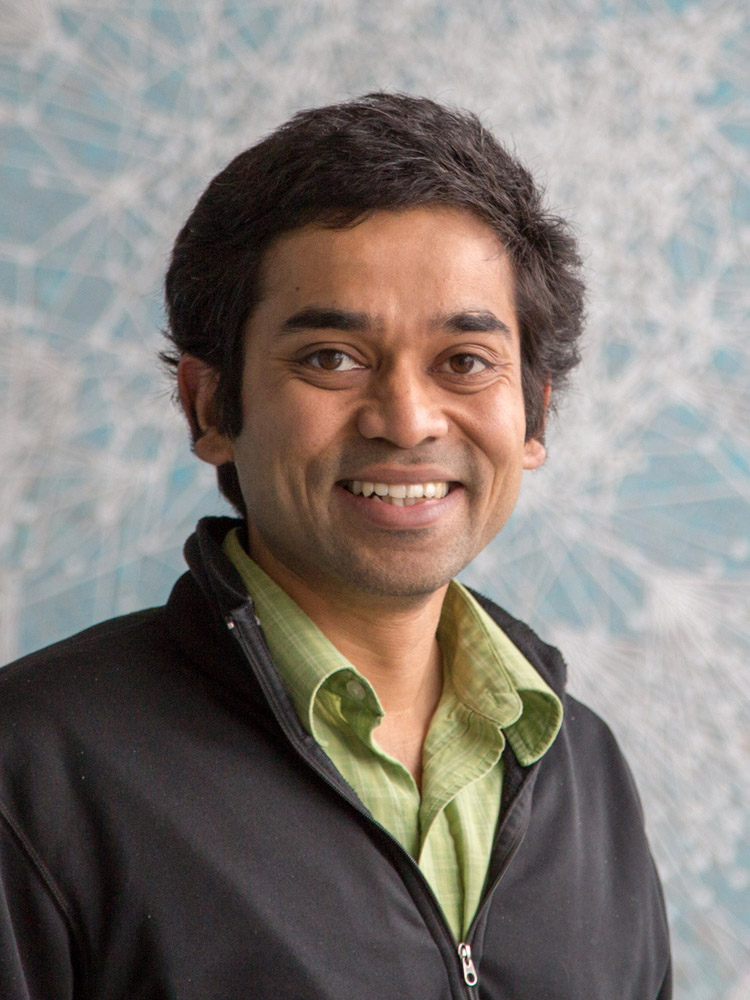}}]{Niloy J.~Mitra}
	leads the Smart Geometry Processing group in the Department of Computer Science at University College London. He received his PhD from Stanford University under the guidance of Leonidas Guibas. His research interests include shape analysis, creativeAI, and computational design and fabrication. Niloy received the Eurographics Outstanding Technical Contributions Award in 2019, the BCS Roger Needham award in 2015, and the ACM Siggraph Significant New Researcher Award in 2013.
\end{IEEEbiography}

\begin{IEEEbiography}[{\includegraphics[width=1in,height=1.25in,clip,keepaspectratio]{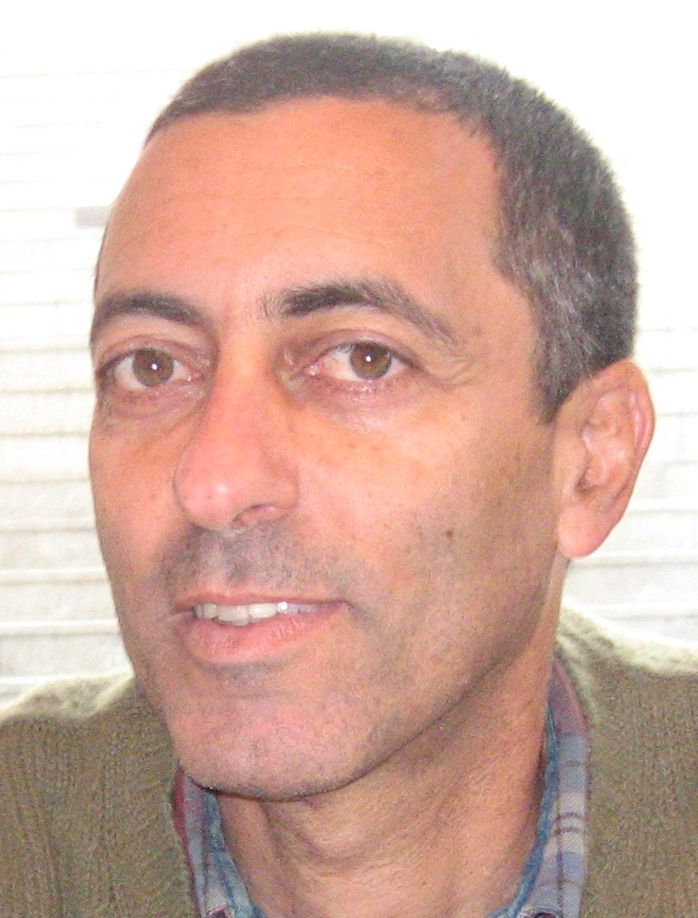}}]{Daniel~Cohen-Or}
	is a Professor in Computer Science. He received his Ph.D. from the State University of New York at Stony Brook in 1991. He was the recipient of Eurographics Outstanding Technical Contributions Award in 2005, ACM SIGGRAPH Computer Graphics Achievement Award in 2018. In 2013 he received The People’s Republic of China Friendship Award. In 2015 he was named a Thomson Reuters Highly Cited Researcher. In 2019 he won Kadar Family Award for Outstanding Research. In 2020 he received Eurographics Distinguished Career Award. His research interests are in Computer Graphics, in particular, synthesis, processing and modeling. 
\end{IEEEbiography}

\begin{IEEEbiography}[{\includegraphics[width=1in,height=1.25in,clip,keepaspectratio]{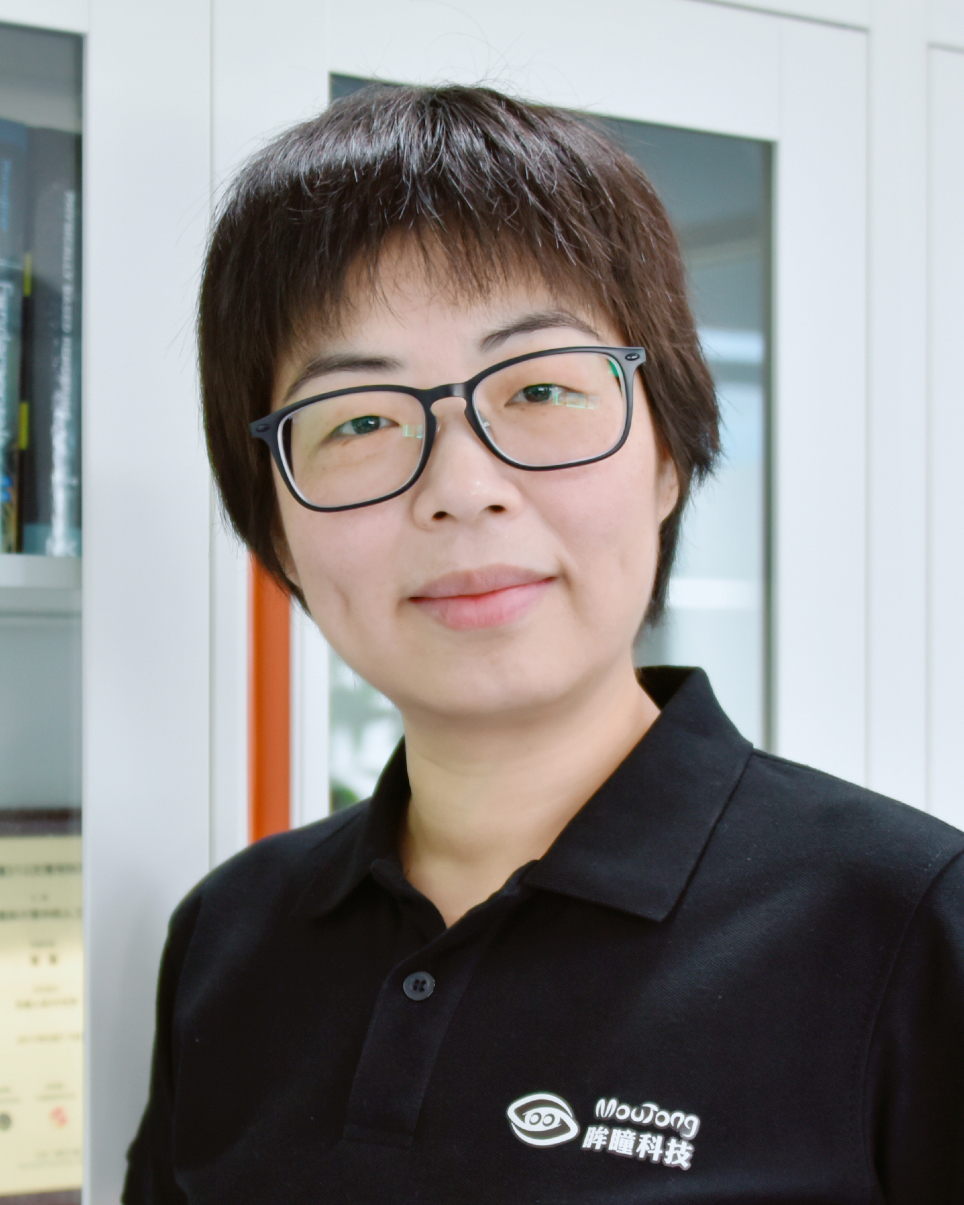}}]{Hui~Huang}
	 is a Distinguished TFA Professor of Shenzhen University, where she directs the Visual Computing Research Center. She received her PhD in Applied Math from The University of British Columbia in 2008. Her research interests span on Computer Graphics, Vision and Visualization. She is currently a Senior Member of IEEE/ACM/CSIG, a Distinguished Member of CCF, and is on the editorial board of ACM Trans. on Graphics, IEEE Trans. on Visualization and Computer Graphics, Computers $\&$ Graphics.
\end{IEEEbiography}

\end{document}